\documentclass[times,paper=a4, fontsize=13pt, letterpaper]{article}

\usepackage{arXiv}

\usepackage{times}
\usepackage{epsfig}
\usepackage{graphicx}
\usepackage{amsmath}
\usepackage{amssymb}
\usepackage{multirow}
\usepackage{color}
\usepackage{wrapfig}
\usepackage{verbatim}
\usepackage[toc,page]{appendix}
\usepackage{enumerate}

\begin{document}

\title{Describing Common Human Visual Actions in Images}


\maketitle




Which common human actions and interactions are recognizable in monocular still images? Which involve objects and/or other people? How many is a person performing at a time? We address these questions by exploring the actions and interactions that are detectable in the images of the MS COCO dataset. We make two main contributions. First, a list of 140 common `visual actions', obtained by analyzing the largest on-line verb lexicon currently available for English (VerbNet) and human sentences used to describe images in MS COCO. Second, a complete set of annotations for those `visual actions', composed of subject-object and associated verb, which we call COCO-a (a for `actions'). COCO-a is larger than existing action datasets in terms of number of actions and instances of these actions, and is unique because it is data-driven, rather than experimenter-biased. Other unique features are that it is exhaustive, and that all subjects and objects are localized. A statistical analysis of the accuracy of our annotations and of each action, interaction and subject-object combination is provided.



\section{Introduction}
Vision, according to Marr, is ``to know what is where by looking.'' This is a felicitous definition, but there is more to scene understanding than `what' and `where': there are also `who', `whom', `when' and `how'. Besides recognizing objects and estimating shape and location, we wish to detect agents, understand their actions and plans, estimate what and whom they are interacting with, reason about cause and effect, predict what will happen next.

The idea that actions are an important component of `scene understanding' in computer vision dates back at least to the '80s~\cite{nagel1988image,nagel1994vision}. In order to detect actions alongside objects the relationships between those objects needs to be discovered. For each action the roles of `subject' (active agent) and `object' (passive - whether thing or person) have to be identified. This information may be expressed as a `semantic network'~\cite{russell1995modern}, which is the first useful output of a vision system for scene understanding\footnote{While there is broad agreement that the knowledge produced by a `scene understanding' algorithm will take the form of a graph, the exact contents and the name of this graph have not yet settled. We will call it semantic network here. Other popular names are `parse network', `knowledge graph', `scene graph'.}. Further steps in in scene understanding include assessing causality and predicting intents and future events. It may be argued that producing a full-fledged semantic network for the entire scene may not be necessary in answering questions about the image, as in the Visual Turing Test~\cite{gemanTuringTest2015}, or in producing output in natural language form. One of the goals of the present study is to ground this debate in data and make the discussion more empirical and less philosophical.\\

\begin{figure}
\begin{minipage}[c]{\textwidth}
\begin{center}
\begin{tabular}{c|c}
\hspace*{-0.6cm}{\footnotesize MS COCO image n.248194} & \hspace*{0.2cm}{\footnotesize MS COCO captions}\\
   \hspace*{-0.2cm}\includegraphics[height=0.24\textwidth]{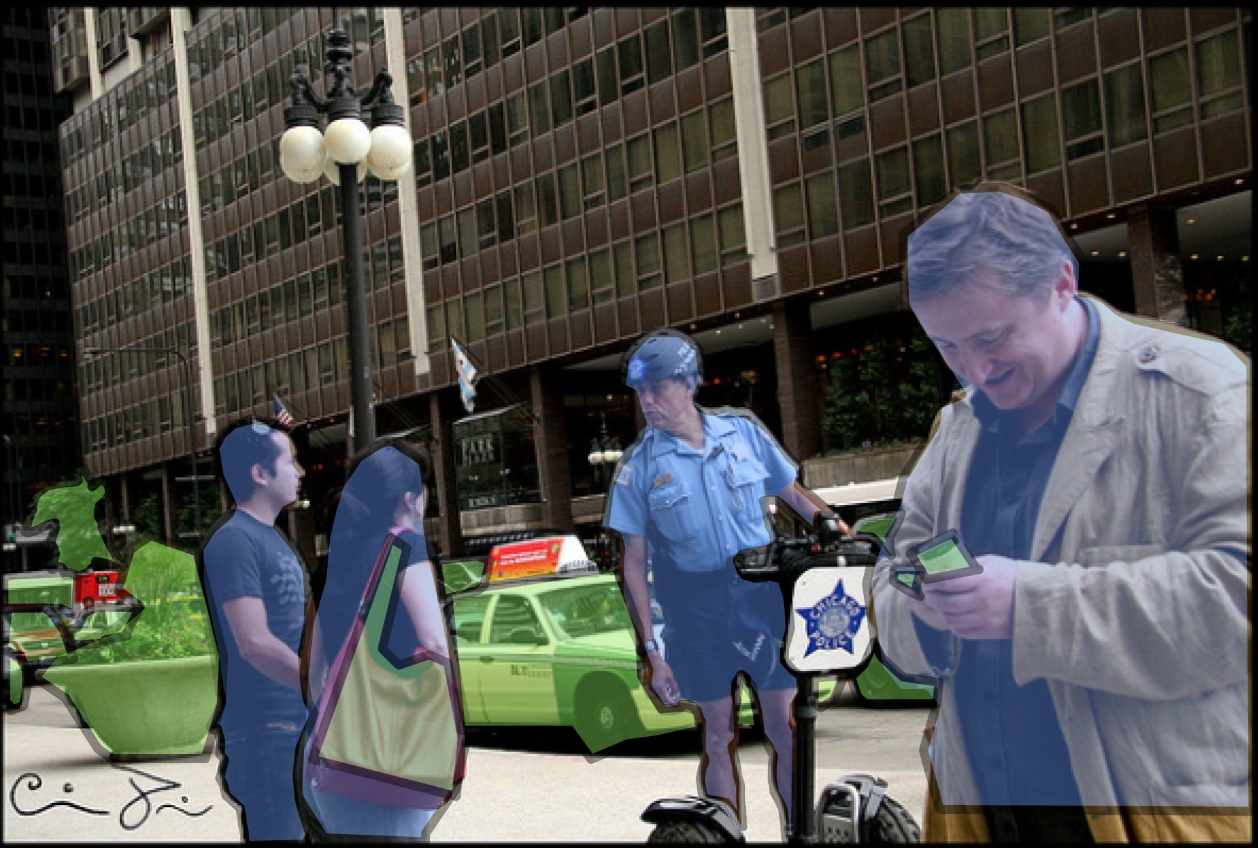} \hspace{0.25in} & \hspace{0.25in}
   \includegraphics[height=0.24\textwidth]{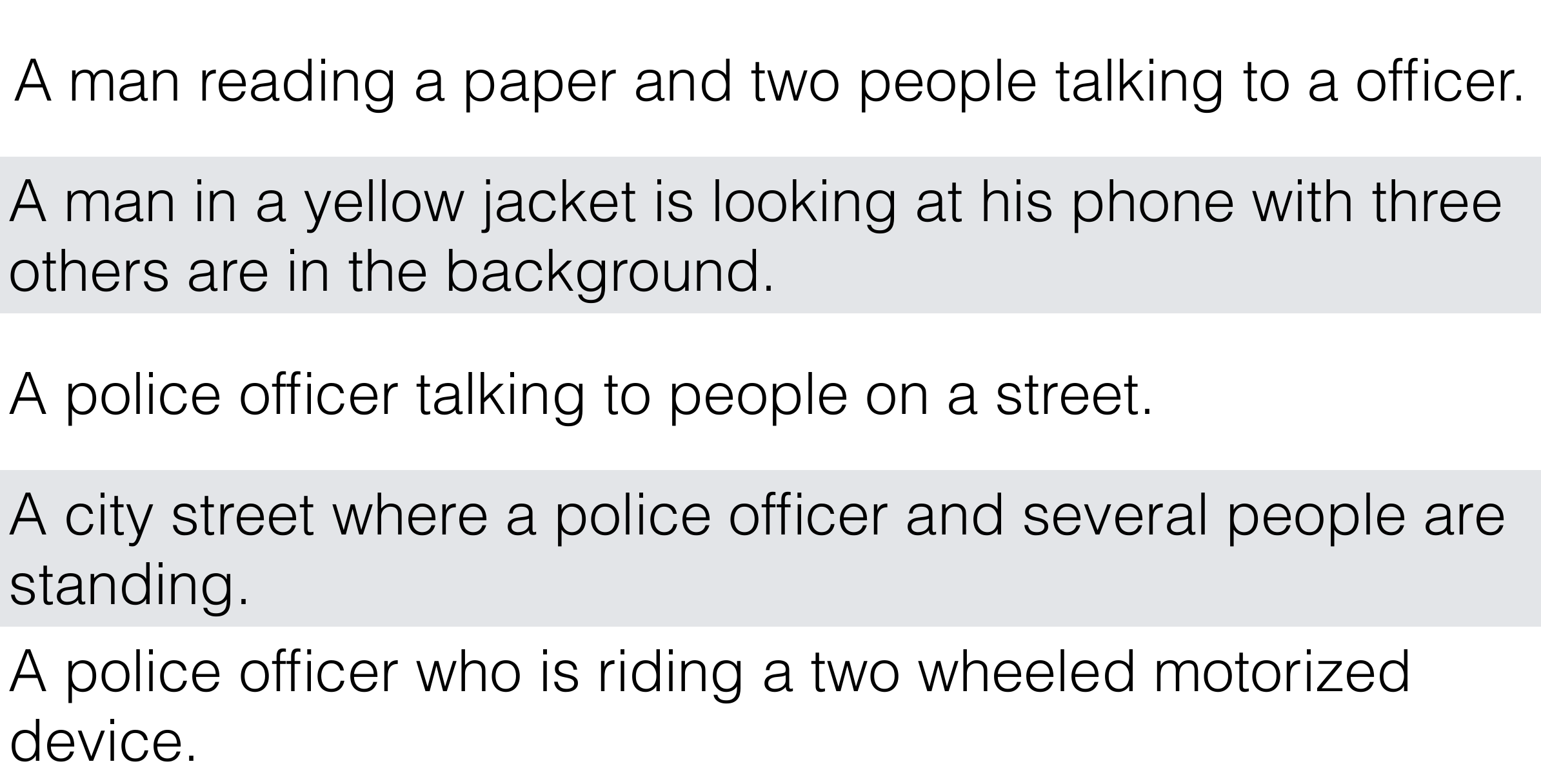}\\
\end{tabular}
\begin{tabular}[t]{c}
\hline
{\footnotesize COCO-a annotations (this paper)}\\
 \hspace*{-0.3cm}\includegraphics[width=\textwidth]{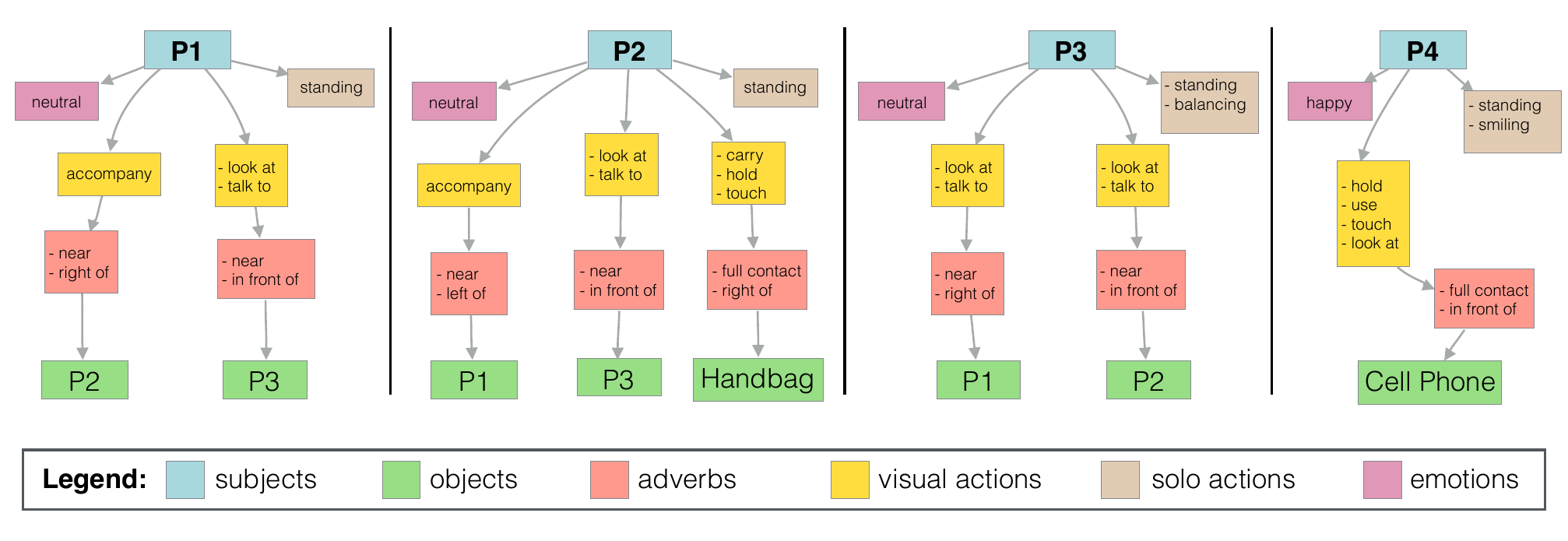}
\end{tabular}
\end{center}
\end{minipage}
\vspace*{-0.4cm}
\caption{{\bf COCO-a annotations}. (Top) MS COCO image and corresponding MS COCO captions. (Bottom) COCO-a annotations. Each person (denoted by P1--P4, left to right in the image) is in turn a subject (blue) and an object (green). Annotations are organized by subject. Each subject and each subject-object pair is associated to states and actions. Each action is associated to one of the $140$ visual actions in our dataset.}
\label{fig:intro_cocoa}
\end{figure}

Three main challenges face us in approaching scene understanding. (1) Deciding the nature of the representation that needs to be produced (e.g. there is still disagreement on whether actions should be viewed as arcs or nodes in the semantic network). (2) Designing algorithms that will analyze the image and produce the desired representation. (3) Learning -- most of the algorithms that are involved have a considerable number of free parameters. In the way of each one of these steps is a dearth of annotated data. 

The ideal dataset to guide our next steps has four desiderata: (a) it is {\em representative} of the pictures we collect every day; (b) it is {\em richly and accurately annotated} with the type of information we would like our systems to know about; (c) it is {\em not biased} by a particular approach to scene understanding, rather it is collected and annotated independently of any specific computational approach; (d) it is {\em large}, containing sufficient data to train the large numbers of parameters that are present in today's algorithms. Current datasets do not measure up to one or more of these criteria. Our goal is to fill this gap.
In the present study we focus on actions that may be detected from single images (rather than video). We explore the visual actions that are present in the recently collected MS COCO image dataset~\cite{lin2014microsoft}. The MS COCO dataset is large, finely annotated and focussed on 81 commonly occurring objects and their typical surroundings. 

By studying the visual actions in MS COCO we make two main contributions:\\
{\bf 1.} An {\bf unbiased method for estimating actions}, where the data tells us which actions occur, rather than starting from an arbitrary list of actions and collecting images that represent them. We are thus able to explore the type, number and frequency of the actions that occur in common images. The outcome of this analysis is {\bf Visual VerbNet (VVN)} listing the 140 common actions that are visually detectable in images.\\
{\bf 2.} A {\bf large and well annotated dataset of actions} on the current best image dataset for visual recognition, with rich annotations including not only all the actions performed by each person in the dataset, but also all the people and objects that are involved in each action, subject's posture and emotion, and high level visual cues such as mutual position and distance (Figure~\ref{fig:intro_cocoa}).


\section{Previous Work}
Human action recognition has been an important research topic in Computer Vision since the late 80's, and was mainly based on motion/video datasets.
Nagel and his collaborators analyzed the German language to detect verbs that refer to actions in urban traffic scenes. They found 119 verbs referring to 67 distinct actions~\cite{von1988formalismus,koller1991algorithmic}, a complete description of actions in a well-defined environment of practical relevance.

Early work on human action detection focussed on detecting actions as spatio-temporal patterns~\cite{polana1992recognition,rohr1994towards} and was unconcerned with the position of the interaction of agents with objects. Datasets collected in the early 2000s reflect this interest. 
A popular example is the KTH dataset~\cite{schuldt2004recognizing} containing video of people performing 6 actions (no interaction with objects and other people). Laptev and collaborators~\cite{laptev2008learning} collected the {\em Hollywood} dataset culling video from commercial movies, thus  removing experimenter bias from acting and filming. They selected 12 classes of human actions and annotated their dataset accordingly. This is a pre-segmented video dataset containing 3669 video clips for 20 hours of video in total. The agents, scenes and objects are not annotated.

Exploring actions in still images~\cite{guo2014survey} is very valuable given the prevalence and convenience of still pictures. It  presents additional challenges -- detecting humans, and computing their pose, is more difficult than in video, and the direction of motion is not available making some actions ambiguous (e.g. picking up versus putting down a pen on a desk). State-of-the-art datasets are summarized in Table~\ref{table-datasets}.

\begin{table}
\begin{center}
\resizebox{\textwidth}{!}{
\begin{tabular}{| l |c|c||c|c|c|c|c|}
\hline 
        &        &         & \multicolumn{5}{|c|}{Per Image Statistics}\\
Dataset & Images & Actions & Subjects & Objects & Interactions & Actions & Adverbs \\ 
\hline
\hline
Pascal~\cite{pascal-voc-2012}      & $ 9100 $ & $ 10 $    &   $1$ & $1$   & {\footnotesize x}     & $1$      & {\footnotesize x} \\
Stanford 40~\cite{yao2011human}    & $ 9532 $ & $ 40$     &   $1$ & $1$   & {\footnotesize x}     & $1$      & {\footnotesize x} \\
89 Actions~\cite{le2013exploiting} & $ 2038 $ & $ 89 $    &   $1$ & $1$   & {\footnotesize x}     & $1$      & {\footnotesize x} \\
TUHOI~\cite{le2014tuhoi}           & $ 10805$ & $ 2974 $  & $1.8$ &  -    & {\footnotesize x}     & $4.8$    & {\footnotesize x} \\
\textbf{Our work}                  & $ \sim10^4 $ & $140$ & $2.2$ & $5.2$ & $5.8$ & $11.1$ & $9.6$\\
\hline
\end{tabular}
}
\end{center}
\vspace*{-0.5cm}
\caption{\textbf{State of the art datasets in single-frame action recognition. } We indicate with `{\footnotesize x}' quantities that are not annotated, with `-' statistics the are not reported. The meaning of Interactions and Adverbs is explained in Section~\ref{sec:dataset}. }
\label{table-datasets}
\end{table}

Everingham and collaborators annotated the PASCAL dataset with 10 actions~\cite{pascal-voc-2012} as a part of the PASCAL-VOC competition. The dataset contains images from multiple sources. The dataset is annotated for objects, and contains a point location for human bodies. Fei-Fei and collaborators collected the {\em Stanford 40 Action Dataset} with images of humans performing 40 actions~\cite{yao2011human}. All images were obtained from Google, Bing, and Flickr. The person performing the action is identified by a bounding box, but objects are not localized. There are $9532$ images in total and between 180 and 300 images per action class. Le et al. in their {\em 89 Actions Dataset}~\cite{le2013exploiting} selected all the images in PASCAL representing a human action and assembled a dataset of $2038$ images, which they manually annotated with a verb. The dataset contains 19 objects and 36 verbs, which are combined to form 89 actions. 

MS COCO has been annotated with ten captions per image~\cite{lin2014microsoft}, which provides information on actions. These annotations have many good properties: they are data-driven and unbiased; easy and inexpensive to collect; intuitive and familiar for human interpretation. However, from the point of view of training algorithms for action recognition there are significant drawbacks: captions don't specify where things are in the image; captions focus typically on one action, a very incomplete description of the image; natural language is ambiguous and still difficult to analyze automatically. For these reasons the MS COCO captions are not sufficient to inform research on action recognition.\\

The closest work to our own, at least in spirit, is a dataset called TUHOI~\cite{le2014tuhoi}. It is based on the annotations in ImageNet~\cite{deng2009imagenet} and adds annotations to localize actions in images. However, verbs are free-typed by the annotators, which does not guarantee that actions are visually discriminable, introduces many ambiguities (such as synonyms) and does not control the specificity of the verbs -- more on this in the next section.

In the present paper we make a number of steps forward. First, we derive actions from the data rather than imposing a pre-defined set of actions. Second, we collect data in the form of semantic networks, in which active entities and all the objects they are interacting with are represented as connected nodes. Each agent-object pair is labelled with the set of relevant actions; each agent is also labelled with the `solo' actions such as posture and motion. Meta-data such as emotional state of the agent, relative location and distance at which interactions occur is also recorded. The advantages of this representation over natural language captions can be seen in Figure~\ref{fig:intro_cocoa}.

\section{Framework}
It is important to keep the distinction straight between `verbs' and `actions'. Verbs are words and actions are states and events. According to the dictionary, a verb is ``a word used to describe an action, state, or occurrence''. By contrast, an action is ``the fact or process of doing something''. Thus {\em verbs} are words that are used to denote {\em actions}. Unfortunately, the correspondence between verbs and actions is not one-to-one. For example, the verb {\it spread} may denote the action of spreading jam on a toast using a knife, or may describe the action carried out by a group of people who part ways simultaneously. Same word, different actions. Conversely, \textit{to spread} (in the culinary sense) becomes \textit{to butter} when what is being spread is butter. Two words for the same action. Furthermore, some actions may be denoted by a single word, \textit{surf} or \textit{golf}, and others may require a few words, \textit{play tennis} and \textit{ride a bicycle}. For simplicity we will call `verb' all the expressions that describe actions, whether single or multi-worded.
 
Actions are not equal in length and complexity. It has been pointed out that one may distinguish between `movemes', `actions', and `activities'~\cite{bregler97,anderson2014toward} depending on structure, complexity, and duration. For example: \textit{reach} is a moveme (a brief target-directed ballistic motion), \textit{drink from a glass} is an action (a concatenation of movemes: reach the glass, grasp its stem, lift the glass to the lips etc.), while \textit{dine} is an activity (a stochastic concatenation of actions taking place over a stretch of time). Here we do not distinguish between movemes, actions and activities because in still images the extent in time and complexity is not directly observable.

We call `\textbf{visual action}' an action, state or occurrence that has a unique and unambiguous visual connotation, making it detectable and classifiable; i.e., \textit{lay down} is a visual action, while \textit{relax} is not. A visual action may be discriminable only from video data, `multi-frame visual action' such as \textit{open} and \textit{close}, or from monocular still images, `single-frame visual action' (simply `visual action' throughout the rest of this paper), such as \textit{stand}, \textit{eat} and \textit{play tennis}.

In order to label visual actions we will use the verbs that come readily to mind to a native English speaker, a concept akin to {\em entry-level categorization} for objects~\cite{palmer99}. Based on this criterion sometimes we prefer more general visual actions (e.g. \textit{play tennis}) rather than the sports domain specific ones such as \textit{volley} or \textit{serve}, and \textit{drink} rather than more specific `movemes' such as \textit{lift a glass to the lips}), other times more specific ones (e.g. \textit{shaking hands} instead of more generally \textit{greet}).

While taxonomization has been adopted as an adequate means of organizing object categories (e.g. animal $\rightarrow$ mammal $\rightarrow$ dog $\rightarrow$ dalmatian), and shallow taxonomies are indeed available for verbs in VerbNet~\cite{kipper2008large}, we are not interested in fine-grained categorization for the time being and do not believe that MS COCO would support it either. Thus, there are no taxonomies in our set of visual actions.

\section{Dataset collection}
\label{sec:dataset}

\begin{figure}
\centering
\begin{minipage}[c]{\textwidth}
\includegraphics[width=\textwidth]{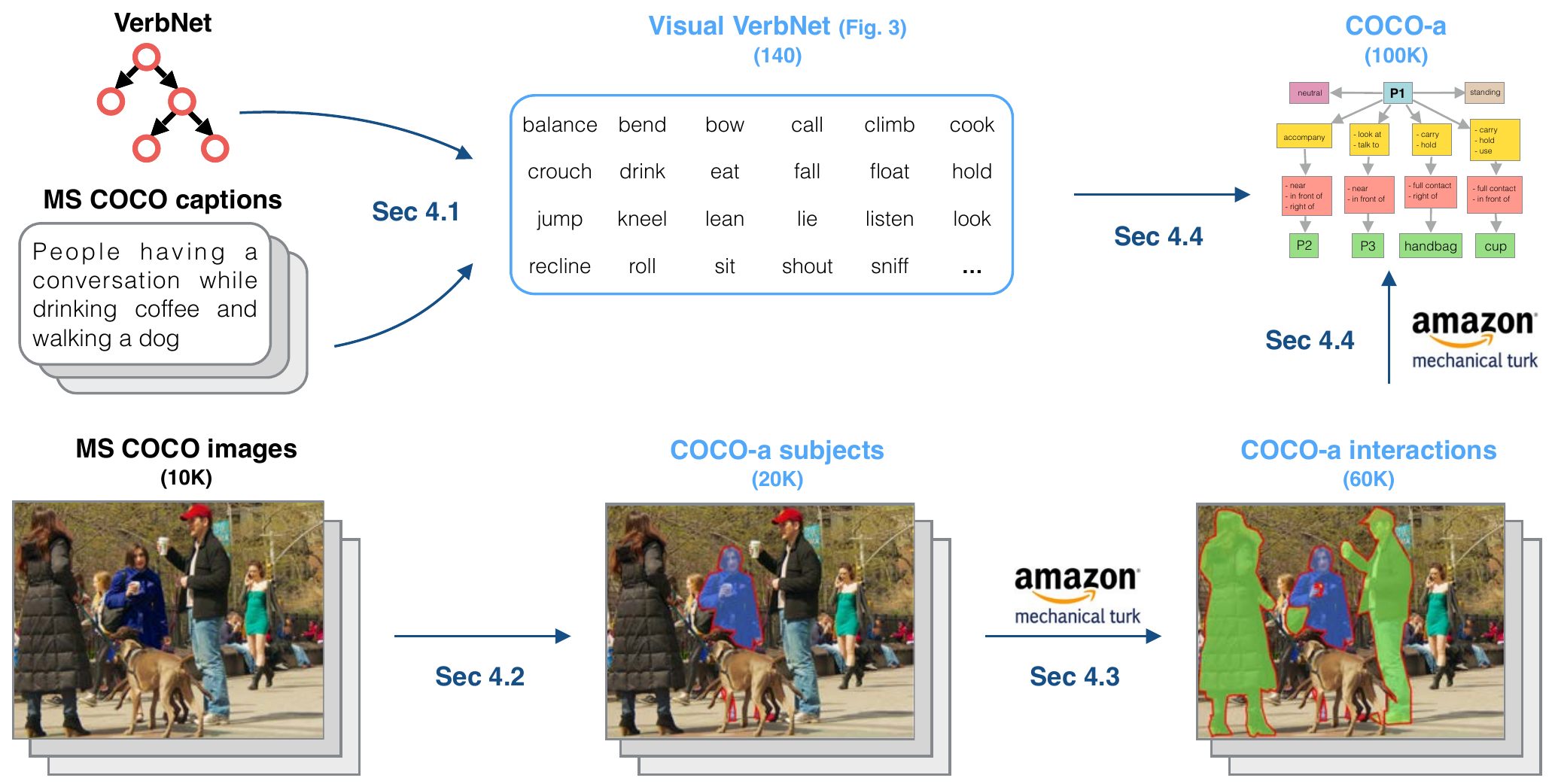}
\end{minipage}
\caption{\textbf{Steps in the collection of COCO-a.} From VerbNet and MS COCO captions we extracted a list of visual actions. All the persons that are annotated in the MS COCO images were considered as potential `subjects' of actions, and paid workers annotated all the objects they interact with, and assigned the corresponding visual actions. Titles in light blue indicate the components of the dataset. Numbers 4.X indicate subsections where each step is described. MS COCO image n.118697 is used in the Figure.}
\label{fig:dataset-steps}
\end{figure}

Our goal is to collect an unbiased dataset with a large amount of meaningful and detectable interactions involving human agents as subjects. We put together a process, exemplified in Figure~\ref{fig:dataset-steps}, consisting of four steps: (Section~\ref{sec:vvn}) Obtain the list of common visual actions that are observed in everyday images. (Section~\ref{sec:img_subj_sel}) Identify the people who are carrying out actions (the subjects). (Section~\ref{sec:interactions_anno}) For each subject identify the objects that he/she is interacting with. (Section~\ref{sec:predicate_annotation}) For each subject-object pair identify the relevant actions.
\vspace*{-0.1cm}
\subsection{Visual VerbNet}
\label{sec:vvn}
\begin{figure}
\begin{minipage}[c]{\textwidth}
\begin{tabular}{cc}
\hspace*{-0.3cm}\resizebox{0.6\textwidth}{!}{
\begin{tabular}[b]{c c c c c c c c}
accompany       & chew        & exchange  & jump           & pay             & punch       & sing           & swim           \\
avoid           & clap        & fall      & kick           & perch           & push        & sit            & talk           \\
balance         & clear       & feed      & kill           & pet             & put         & \textbf{skate} & taste          \\
bend (pose)     & climb       & fight     & kiss           & photograph      & reach       & \textbf{ski}   & teach          \\
bend (something)& cook        & fill      & kneel          & pinch           & read        & slap           & \textbf{throw} \\
be with         & crouch      & float     & laugh          & play            & recline     & sleep          & tickle         \\
bite            & cry         & fly       & lay            & play baseball   & remove      & smile          & touch          \\
blow            & cut         & follow    & lean           & play basketball & repair      & sniff          & use            \\
bow             & dance       & get       & lick           & play frisbee    & ride        & snowboard      & walk           \\
break           & devour      & give      & lie            & play instrument & roll        & spill          & wash           \\
brush           & dine        & groan     & lift           & play soccer     & row         & spray          & wear           \\
build           & disassemble & groom     & \textbf{light} & play tennis     & run         & spread         & whistle        \\
bump            & draw        & hang      & listen         & poke            & sail        & squat          & wink           \\
call            & dress       & help      & look           & pose            & separate    & squeeze        & write          \\
caress          & drink       & hit       & massage        & pour            & shake hands & stand          &                \\
carry           & drive       & hold      & meet           & precede         & shout       & steal          &                \\
\textbf{catch}  & drop        & hug       & mix            & prepare         & show        & straddle       &                \\
chase           & eat         & hunt      & \textbf{paint} & pull            & signal      & surf           &                \\              
\end{tabular} } & 
\includegraphics[width=0.4\textwidth]{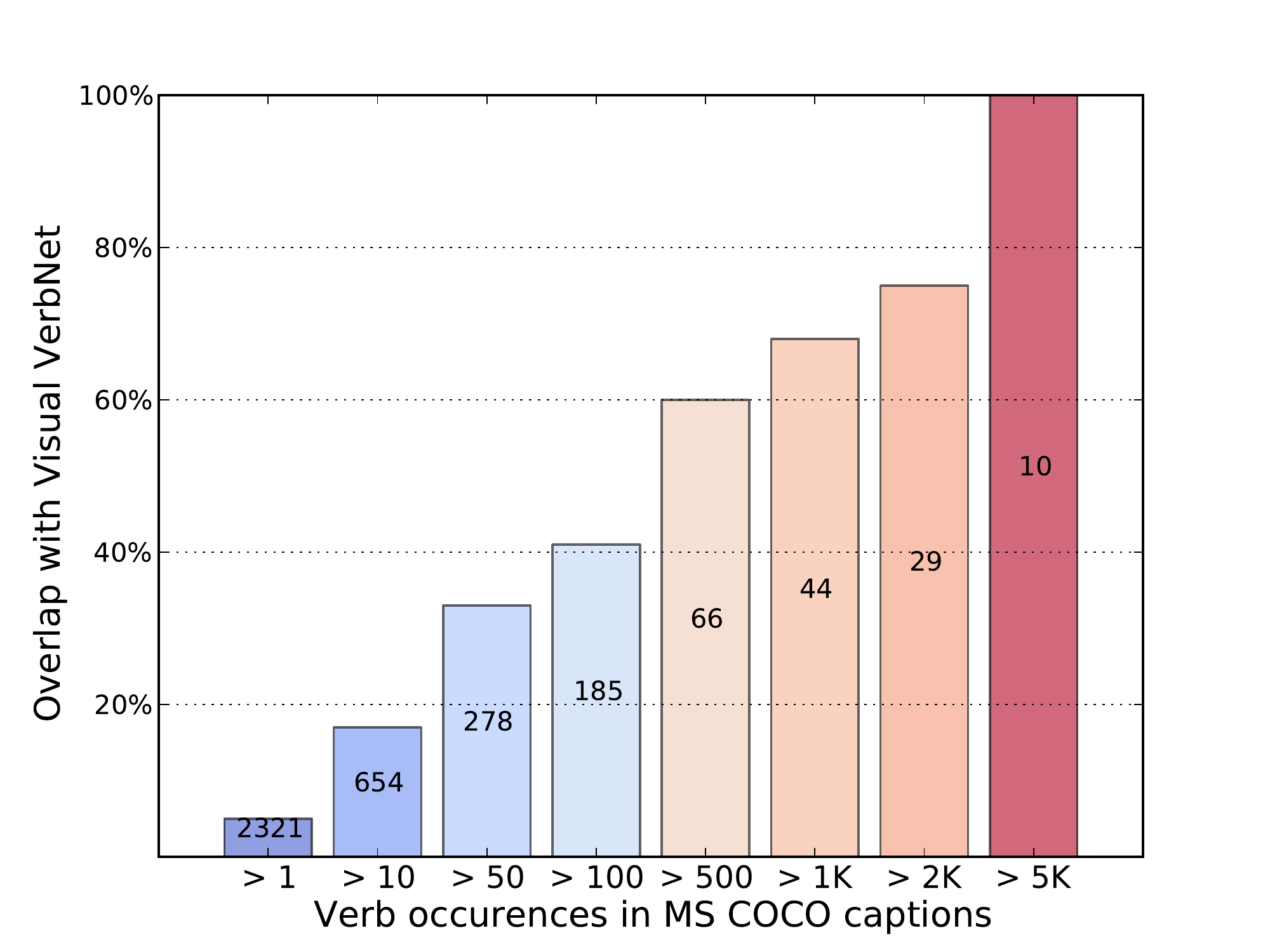}\\
\end{tabular}
\\
\resizebox{\textwidth}{!}{
\begin{tabular}[b]{| c c c c | c c c | c c | c c c c | c |}
\hline
\hline
\multicolumn{4}{|c|}{Not Visual Actions} & \multicolumn{3}{|c|}{Mutli-frame Visual Actions} & \multicolumn{2}{|c|}{Single-Frame Visual Actions} & \multicolumn{4}{|c|}{Synonyms} & Domain-Specific\\
\hline
attempt   & engage  & practice & share   & approach & leave  & start & cover    & tie  & adjust  & gather  & paddle & stare & bowl  \\
board     & enjoy   & prop     & stick   & block    & miss   & step  & face     & wrap & attach  & grab    & pass   & stuff & grind \\
celebrate & extend  & race     & stretch & close    & move   & stop  & line     &      & color   & hand    & pick   & take  & park  \\
check     & feature & reflect  & top     & come     & open   & turn  & load     &      & crowd   & lead    & place  & toss  & pitch \\
compete   & include & relax    & travel  & cross    & raise  &       & slide    &      & display & leap    & say    & watch & set   \\
contain   & learn   & rest     & try     & enter    & return &       & sprinkle &      & dock    & make    & see    & wave  & swing \\
decorate  & live    & seem     & wait    & flip     & seat   &       & stack    &      & handle  & mount   & slice  &       & tow   \\
double    & perform & shape    &         & head     & shake  &       & surround &      & fix     & observe & speak  &       &       \\
\hline  
\end{tabular} }
\end{minipage}
\caption{\textbf{Visual VerbNet (VVN).} (Top-Left) List of 140 visual actions that constitute VVN -- bold ones were added after the comparison with MS COCO captions. (Top-Right) Overlap between the verbs in VVN and in the captions of MS COCO. There is $60\%$ overlap for the $66$ verbs (of the total $2321$ in MS COCO captions) with more than 500 occurrences. (Bottom) Verbs with $>100$ occurrences in the MS COCO captions not contained in VVN, organized in categories. The $10$ single frame visual actions might have been  included in VVN but did not entirely meet our criteria.
}
\label{fig:verbnet-coco-comparison}
\end{figure}

To obtain the list of the entry-level visual actions we examined VerbNet~\cite{kipper2008large} (containing $> 8000$ verbs organized in about $300$ classes) and selected all the verbs that refer to visually identifiable actions. Our criteria of selection is that we would expect a 6--8 year old child to be able to easily distinguish visually between them. This criterion led us to group synonyms and quasi-synonyms (\textit{speak} and \textit{talk}, \textit{give} and \textit{hand}, etc.) and to eliminate verbs that were domain-specific (\textit{volley}, \textit{serve}, etc.) or rare (\textit{cover}, \textit{sprinkle}, etc.). To be sure that we were not missing any important actions, we also analyzed the verbs in the captions of the images containing humans in the MS COCO dataset, and discarded verbs not referring to human actions, without a clear visual connotation, or synonyms. This resulted in adding six additional verbs to our list for a total of 140 visual actions, shown in Figure~\ref{fig:verbnet-coco-comparison} (Left). Figure~\ref{fig:verbnet-coco-comparison} (Right) explores the overlap of VVN with the verbs in MS COCO captions. The overlap is high for verbs that have many occurrences, and verbs that appear in the MS COCO captions and not in VVN do not denote a visual action, are synonyms, or refer to actions that are either very domain-specific or highly unusual, as shown in the table in Figure~\ref{fig:verbnet-coco-comparison} (Bottom). The process we followed ensured an unbiased selection of visual actions. 

Furthermore, we asked Amazon Mechanical Turk (AMT) workers for feedback on the completeness of this list and, given their scant response, we believe that VVN is very close to complete and should not need extension unless specific domain action recognition is required. 

\subsection{Image and subject selection}
\label{sec:img_subj_sel}
Different actions usually occur in different environments, so in order to balance the content of our dataset we selected an approximately equal number images of three types of scenes: sports, outdoors and indoors. We also selected images of various complexity, containing single subjects, small groups (2-4 subjects) and crowds ($>$4 subjects). The exact splits can be found in the Appendix.
From these images, all the people whose pixel area is larger than $1600$ pixels are defined as `subjects'. All the people in an image, regardless of size, are still considered as possible objects of an interaction. The result of this preliminary image analysis is an intermediate dataset containing about $2$ subjects per image, indicated as COCO-a subjects in Figure~\ref{fig:dataset-steps}.
\vspace*{-0.2cm}
\subsection{Interactions annotations}
\label{sec:interactions_anno}
\begin{figure}
\centering
\begin{minipage}[c]{\textwidth}
\begin{tabular}{ccc}
\hspace*{-0.3cm}
\raisebox{0.7cm}{\includegraphics[width=0.3\textwidth]{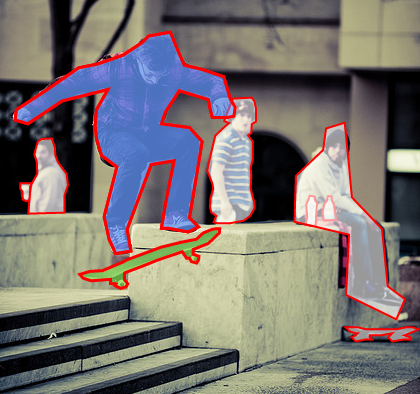}} &
\hspace*{-0.2cm}
\raisebox{0.2cm}{\includegraphics[width=0.45\textwidth]{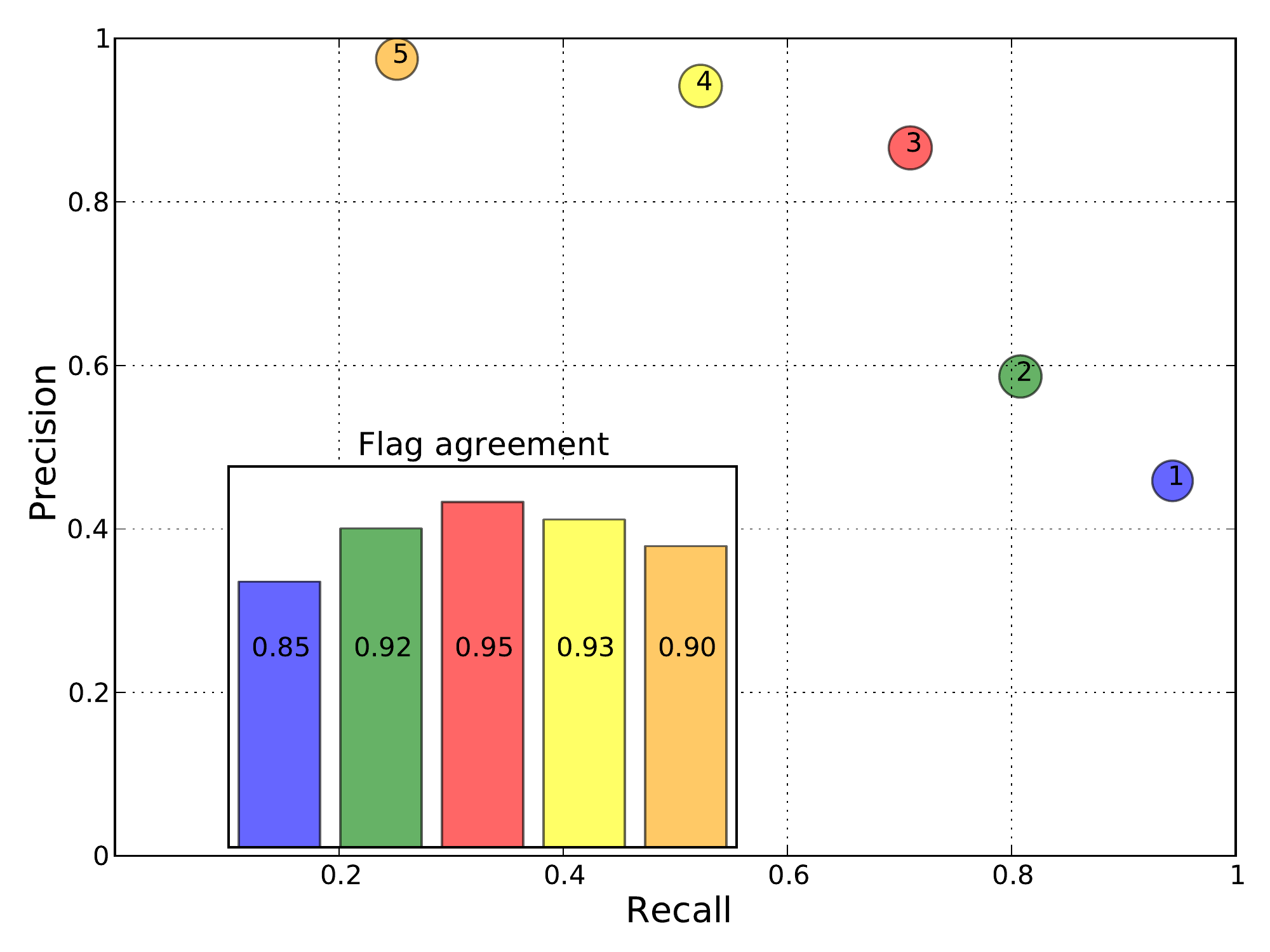}} &
\hspace*{-0.4cm}
 \begin{tabular}[b]{c}
 \includegraphics[width=0.2\textwidth]{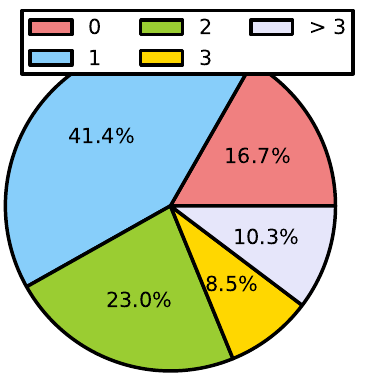}\\
 \includegraphics[width=0.2\textwidth]{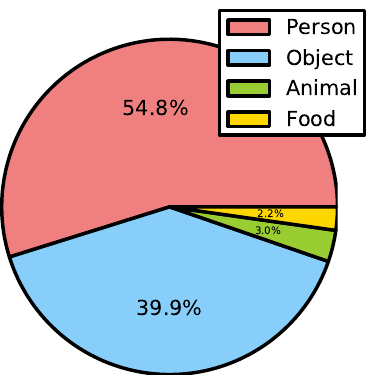}
 \end{tabular}\\
\end{tabular}
\end{minipage}
\vspace*{-0.5cm}
\caption{(Left) \textbf{Interactions GUI.} A snapshot of the GUI presented to the AMT workers. The subject is highlighted in blue, all the possible interacting objects in white, and the provided annotation in green. (Center) \textbf{Quality of the interaction annotations.} Each numbered dot indicates a value of Precision and Recall. The number indicates the number of votes (out of five) that were used to consider the interaction valid. The bar chart shows percentage agreement in discarding subjects that are mostly occluded or invisible. The color refers to the number of votes (same as Precision Recall dots). (Right) \textbf{Statistics.} Distribution of the number of interactions per subject (Top), and category of the interacting objects (Bottom).}
\label{fig:interactions-metrics}
\end{figure}

For each subject, we annotated all the objects that he/she is interacting with. Annotators were presented with images such as in Figure~\ref{fig:interactions-metrics} (Left), containing a highlighted person, the `subject', and asked to either (1) flag the subject if it was mostly occluded or invisible; or (2) click on all the objects he/she is interacting with. Deciding if a person and an object (or other person) are interacting is somewhat subjective, so we asked $5$ workers to analyze each subject and combined their responses. 

In order to assess the quality of the annotations we also collected ground truth from one of the authors for a subset of the images. For each subject-object pair we considered requiring a number of votes ranging from $1$ to $5$. 
We found that three votes yielded the best trade-off between Precision and Recall and the highest flag agreement against our ground truth as shown in Figure~\ref{fig:interactions-metrics} (Center).

After discarding the flagged subjects and consolidating the annotations we obtained an average of $5.8$ interactions per image, which constitute the COCO-a interactions dataset.
As shown in Figure~\ref{fig:interactions-metrics} (Right) about 1/5 of subjects has only `solo' actions (0 objects, red), 2/5 is involved in a single object interaction (1 object, blue), and 2/5 interact with two or more objects (Figure~\ref{fig:intro_cocoa} shows examples of subjects interacting with two and three objects). Figure~\ref{fig:interactions-metrics} (Bottom-Right) suggests that our dataset is human-centric, since more than half of the interactions happen with other people. 

\subsection{Visual Actions annotations}
\label{sec:predicate_annotation}

In the final step of our process we labelled all the subject-object interactions in the COCO-a interactions dataset with the visual actions in VVN. Workers were presented with a GUI containing a single interaction, visualized as in Figure~\ref{fig:interactions-metrics} (Left), and asked to select all the visual actions describing it. In order to keep the collection interface simple, we divided visual actions into 
8 groups -- `\textit{posture/motion}', `\textit{solo actions}', `\textit{contact actions}', `\textit{actions with objects}', `\textit{social actions}', `\textit{nutrition actions}', `\textit{communication actions}', `\textit{perception actions}'. This was based on two simple rules: (a) actions in the same group share some important property, e.g. being performed solo, with objects, with people, or indifferently with people and objects, or being an action of posture; (b) actions in the same group tend to be mutually exclusive. Furthermore, we included in our study $3$ `adverb' categories: `\textit{emotion}' of the subject\footnote{There has been disagreement on the fact that humans might have basic discrete emotions \cite{ortony1990s,du2014compound}. However, we adopt Ekman's $6$ basic emotions~\cite{ekman1992argument} for this study as we are interested in a high level description of subject's emotional state.}, `\textit{location}' and `\textit{relative distance}' of the object with respect to the subject.

This allowed us to obtain a rich set of annotations for all the actions that a subject is performing which completely describe his/her state, a property that is novel with respect to existing datasets and favours the construction of semantic networks centred on the subject. 

We asked three annotators to select all the visual actions and adverbs that describe each subjet-object interaction pair. In some cases annotators interpreted interactions differently, but still correctly. Therefore, we decided to return all the visual actions collected for each interaction along with the value of agreement of the annotators, rather than forcing a deterministic, but arbitrary, ground truth. Depending on the application that will be using our data it will be possible to consider visual actions on which all the annotators agree or only a subset of them. 
The average number of visual action annotations provided per image for an agreement of 1, 2 or all 3 annotators is respectively $19.2$, $11.1$, and $6.1$. This constitutes the content of the COCO-a dataset in its final form.
\subsection{Analysis}
\label{sec:analysis}

Figure~\ref{fig:intro_cocoa} allows a first qualitative analysis of the COCO-a dataset. Compared with MS COCO captions, COCO-a annotations contain additional information by providing: (a) a complete account of all the subjects, objects and actions contained in an image; (b) an unambiguous and machine-friendly form; (c) the specific localization in the image for each subject and object. Statistics of the information that the COCO-a dataset annotations capture and convey for each image is summarized in Table~\ref{table-datasets}\footnote{All Tables, Figures and statistics presented here were computed on a subset of $2500$ images available at the time of writing, and using the agreement of two out of three workers on the `visual action' annotations.}.

\begin{figure}
\begin{minipage}[c]{\textwidth}
\begin{center}
\resizebox{\textwidth}{!}{
\begin{tabular}{cccc}
   \includegraphics[height=0.2\textwidth]{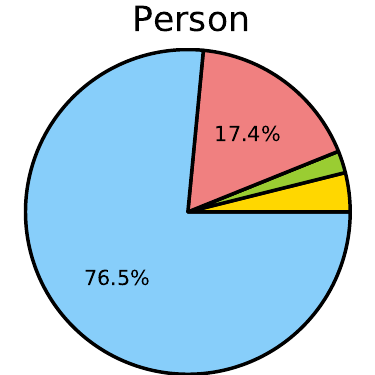}&
   \includegraphics[height=0.2\textwidth]{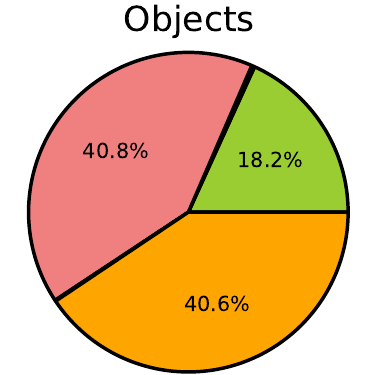}&   
   \includegraphics[height=0.2\textwidth]{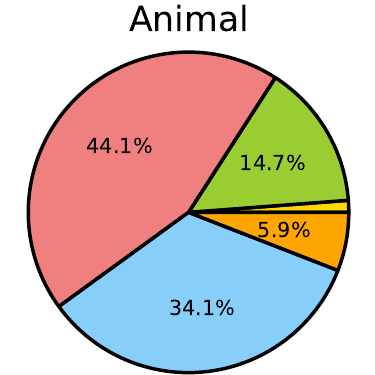}&
   \includegraphics[height=0.2\textwidth]{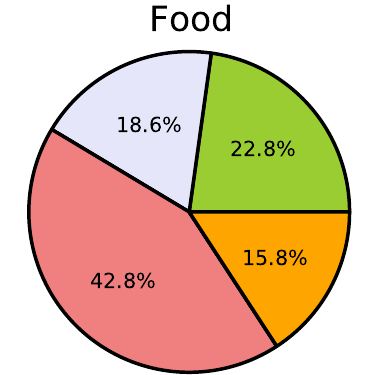}\\
   \multicolumn{4}{c}{\includegraphics[width=0.9\textwidth]{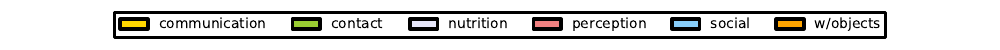}}\\
\end{tabular}
}
\end{center}
\end{minipage}
\vspace*{-0.5cm}
\caption{\textbf{Visual Actions by group. } Fraction of COCO-a visual actions that belong to each of the 6 macro categories when subjects interact with People, Animals, General Objects or Food. We excluded posture and solo actions from this analysis.}
\label{fig:visual_classes_frequencies}
\end{figure}
In Figure~\ref{fig:visual_classes_frequencies} we see the most frequent types of actions carried out when subjects interact with four specific object categories: other people, animals, inanimate objects (such as a handbag or a chair) and food. For interactions with people the visual actions belong mostly to the category `\textit{social}' and `\textit{perception}'. When subjects interact with animals the visual actions are similar to those with people, except there are fewer `\textit{social}' actions and more `\textit{perception}' actions. Person and animal are the only types of objects for which the `communication' visual actions are used at all. When people interact with objects the visual actions used to describe those interactions are mainly from the categories `\textit{with objects}' and `\textit{perception}'. As expected, food items are the only ones that have a good portion of `\textit{nutrition}' visual actions.


\begin{figure}
\begin{minipage}[c]{\textwidth}
\centering
\begin{tabular}{cc}
  & \\
  \hspace*{-0.3cm}\includegraphics[width=0.49\textwidth]{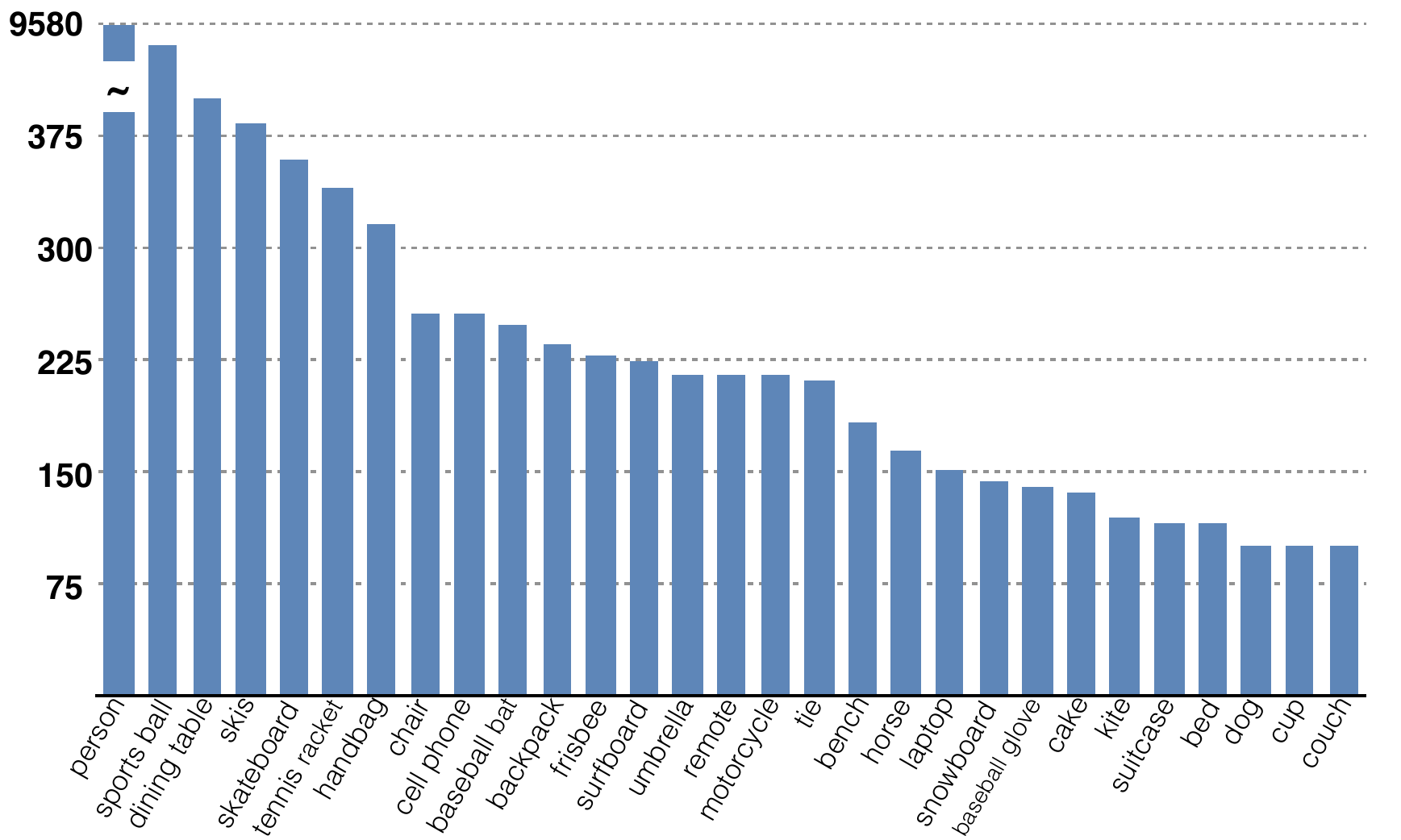}&
  \includegraphics[width=0.49\textwidth]{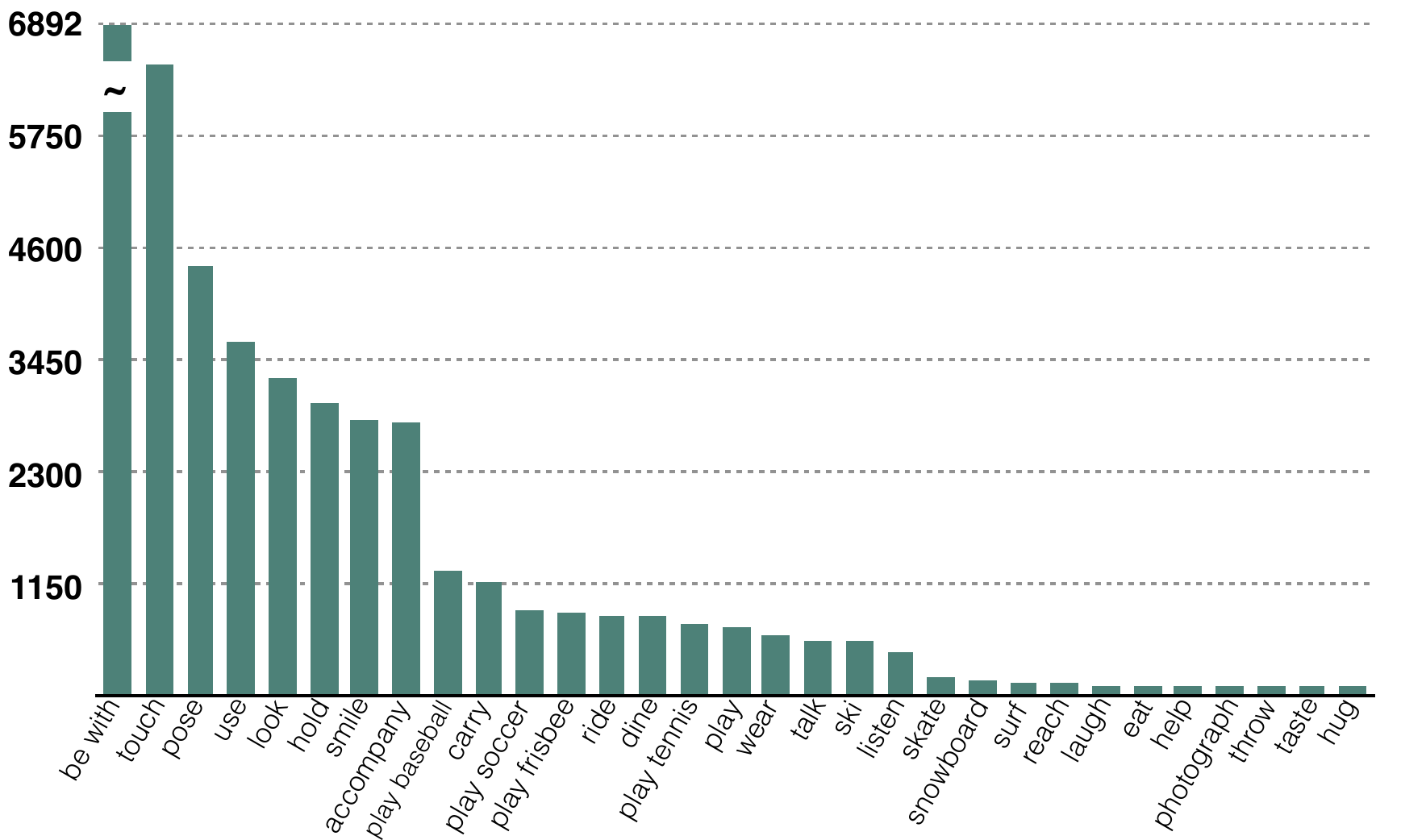}\\
\end{tabular}
\end{minipage}
\vspace*{-0.5cm}
\caption{\textbf{Objects and visual actions.} Count of the most frequent objects that people interact with (Left) and visual actions that people perform (Right) in the COCO-a. We report the 29 objects and 31 visual actions that have more than 100 occurrences. The distributions are long-tailed with a fairly steep slope (Fig.\ref{fig:verbs-list-tail} ).}
\label{fig:object_verb_list}
\end{figure}

Figure~\ref{fig:object_verb_list} (Left) shows the 29 objects with more than 100 interactions in the analyzed images. The human-centric nature of our dataset is confirmed by the fact that the most frequent object of interaction is other persons, an order of magnitude more than the other objects.
Since our dataset contains an equal number of sports, outdoor and indoor scenes, the list of objects is heterogeneous and contains objects that can be found in all environments.\\

In Figure~\ref{fig:object_verb_list} (Right) we list the 31 visual actions that have more than 100 occurrences. It appears that the visual actions list has a very long tail, with $90\%$ of the actions having less than $2000$ occurrences and covering about $27\%$ of the total count of visual actions. This leads to the observation that MS COCO dataset is sufficient for a thorough representation and study of about 20 to 30 visual actions. We are developing methods to bias our image selection process in order to obtain more samples of the actions contained in the tail.

The most frequent visual action in our dataset is `\textit{be with}'. This is a very particular visual action as annotators use it to specify when people belong to the same group. Common images often contain multiple people involved in different group actions, and this annotation can provide insights in learning concepts such as the difference between proximity and interaction -- i.e. two people back to back are probably not part of the same group although spatially close.


\begin{figure}[t!]
\begin{minipage}[c]{\textwidth}
\begin{center}
\begin{tabular}{cc}
   \hspace*{-0.4cm}\includegraphics[width=0.5\textwidth]{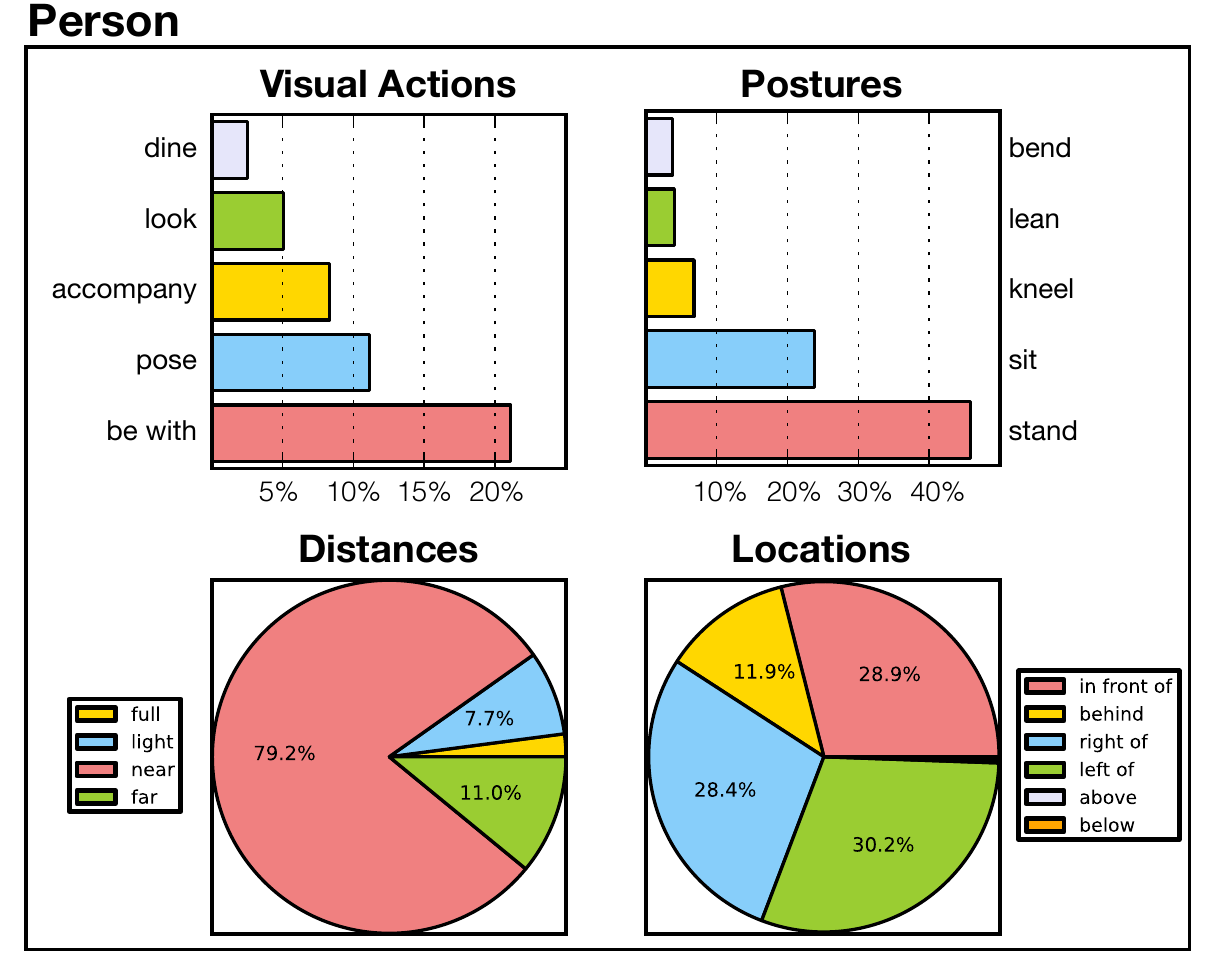}&
   \includegraphics[width=0.5\textwidth]{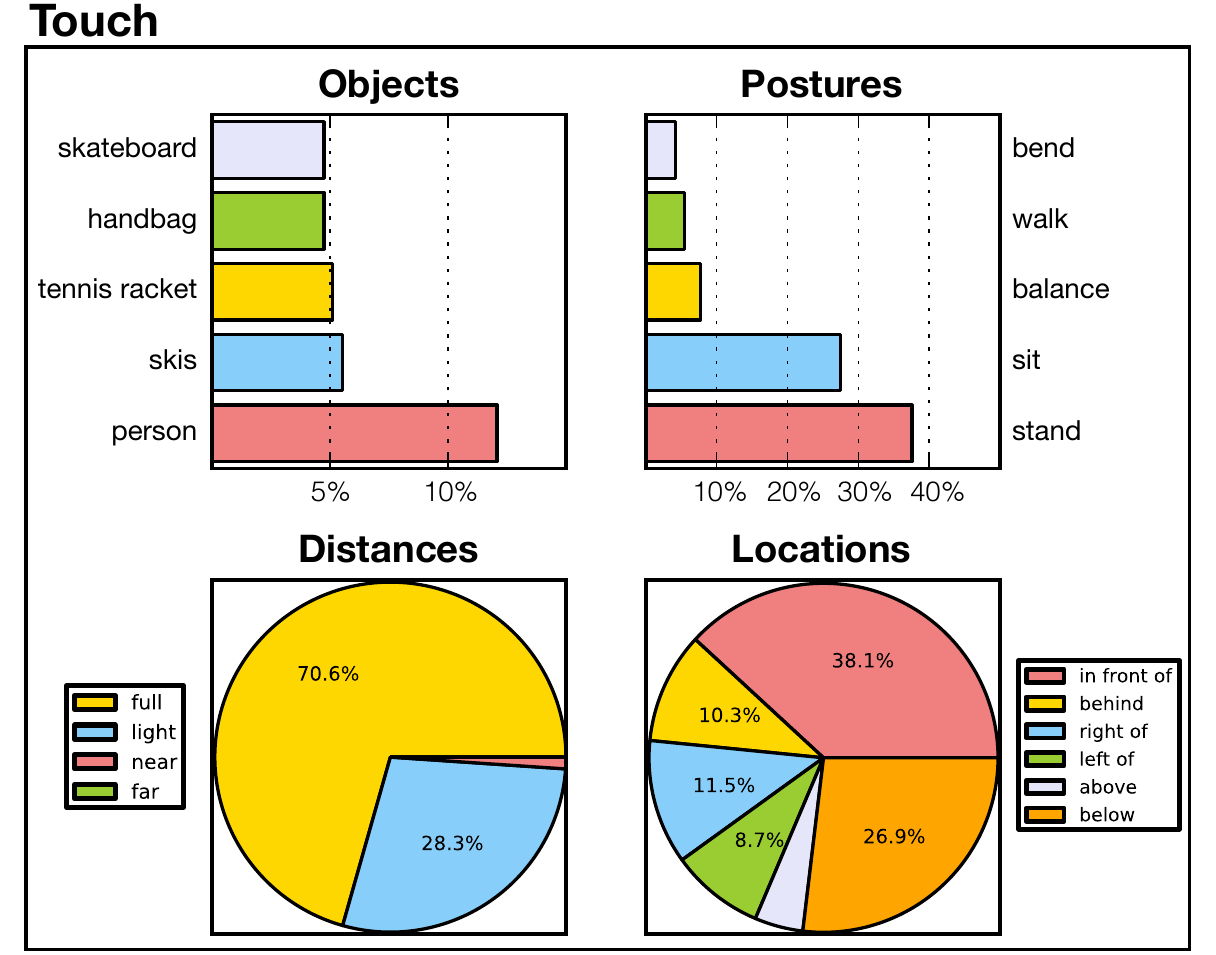}\\
\end{tabular}
\end{center}
\end{minipage}
\vspace*{-0.5cm}
\caption{\textbf{Annotation Analysis. } (Left) Top visual actions, postures, distances and relative locations of person/person interactions. (Right) Objects, postures, distances and locations that are most commonly associated with the visual action `touch'.}
\label{fig:top_verbs_objects}
\end{figure}

The COCO-a dataset provides a rich set of annotations. In Figure~\ref{fig:top_verbs_objects} we provide two examples of the information that can be extracted and explored, for an object and a visual action contained in the dataset. Figure~\ref{fig:top_verbs_objects} (Left) describes interactions between people. We list the most frequent visual actions that people perform together (\textit{be in the same group}, \textit{pose} for pictures, \textit{accompany} each other, etc.), postures that are held (stand, sit, kneel, etc.), distances of interaction (people mainly interact near each other, or from far away if they are playing some sports together) and locations (people are located about equally in front or to each other sides, more rarely behind and almost never above or below each other). A similar analysis can be carried out for the visual action \textit{touch}, Figure~\ref{fig:top_verbs_objects} (Right). The most frequently touched object are other people, sports and wearable items. People touch things mainly when they are standing or sitting (for instance a chair or a table in front of them). As expected, the distribution of locations is very skewed, as people are almost always in full or in light contact when touching an object and never far away from it. The location of objects shows us that people in images usually touch things in front (as comes natural in the action of grasping something) or below of them (such as a chair or bench when sitting).

\begin{figure}[t!]
\begin{center}
\begin{tabular}{ccc}
   \hspace*{0.3cm}\includegraphics[height=0.23\textwidth]{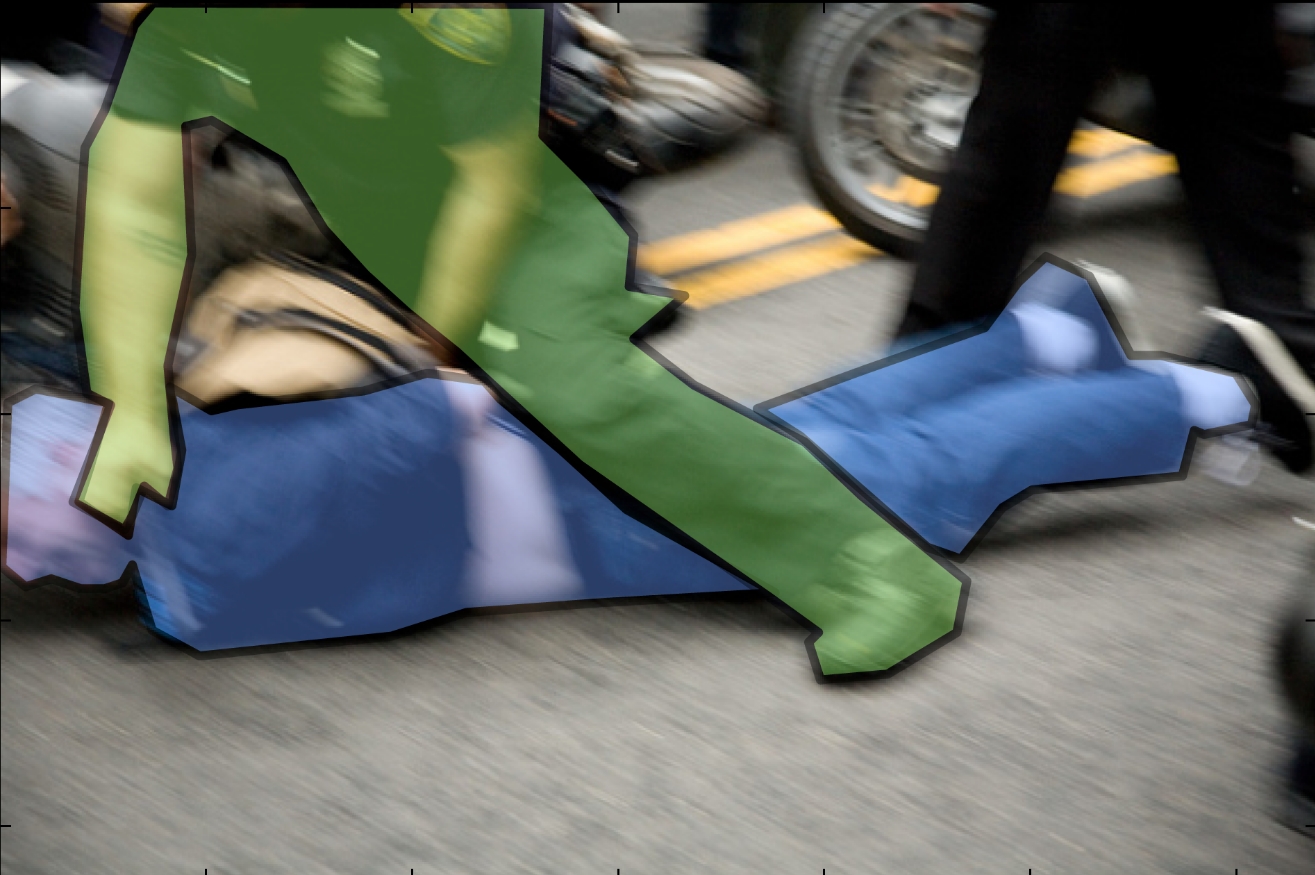}&
   \hspace*{-0.4cm}\includegraphics[height=0.23\textwidth]{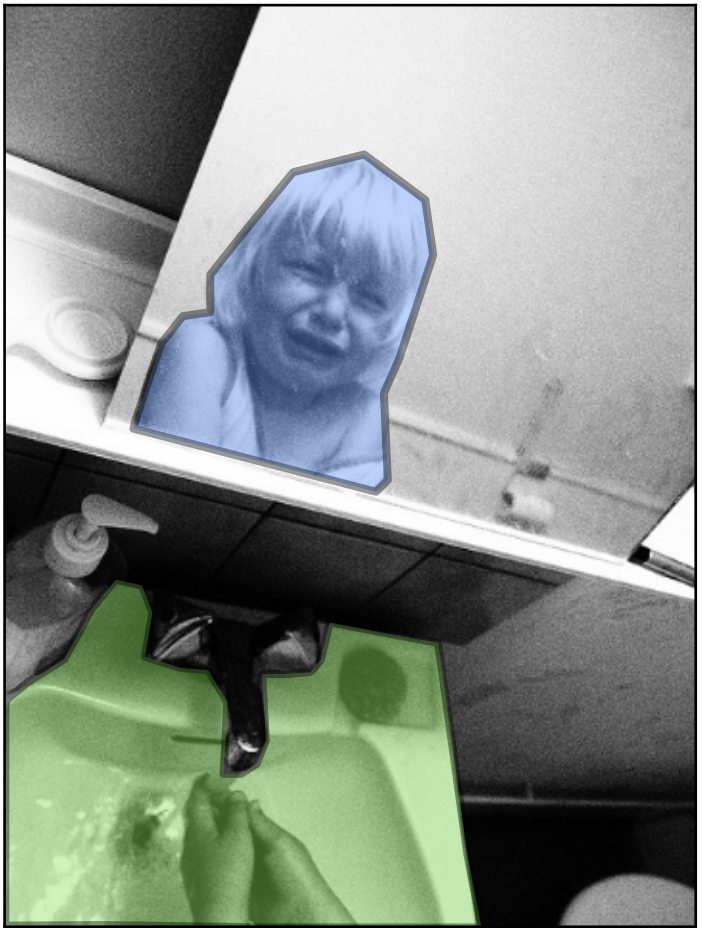}&
   \hspace*{-0.4cm}\includegraphics[height=0.23\textwidth]{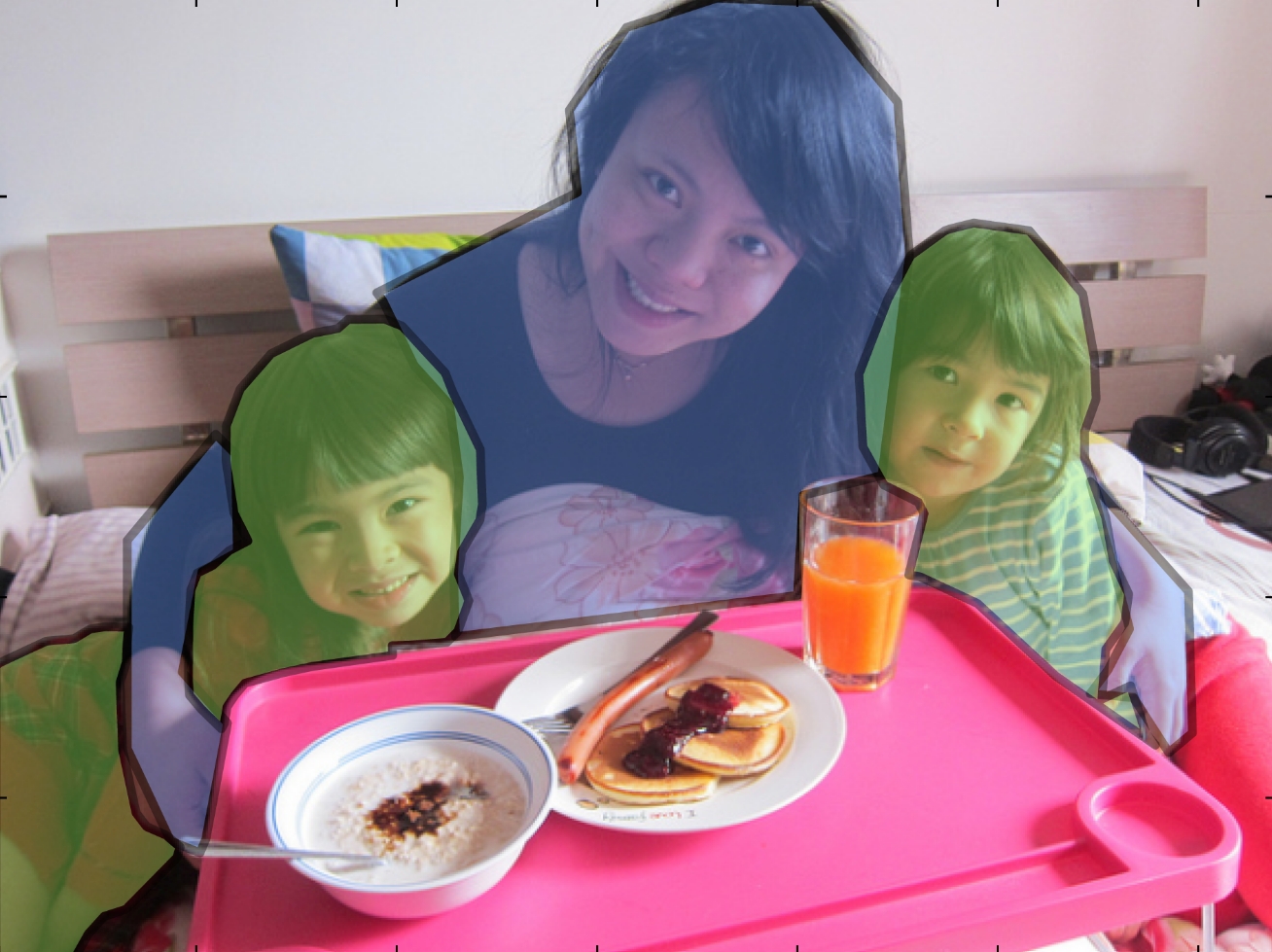}\\
   \hspace*{0.3cm}`fight' + `above' & \hspace*{-0.4cm}`cry' + `sink' & \hspace*{-0.4cm}`pose' + `full contact'\\
\end{tabular}

\begin{tabular}{cc}
   \hspace*{0.2cm}\includegraphics[height=0.23\textwidth]{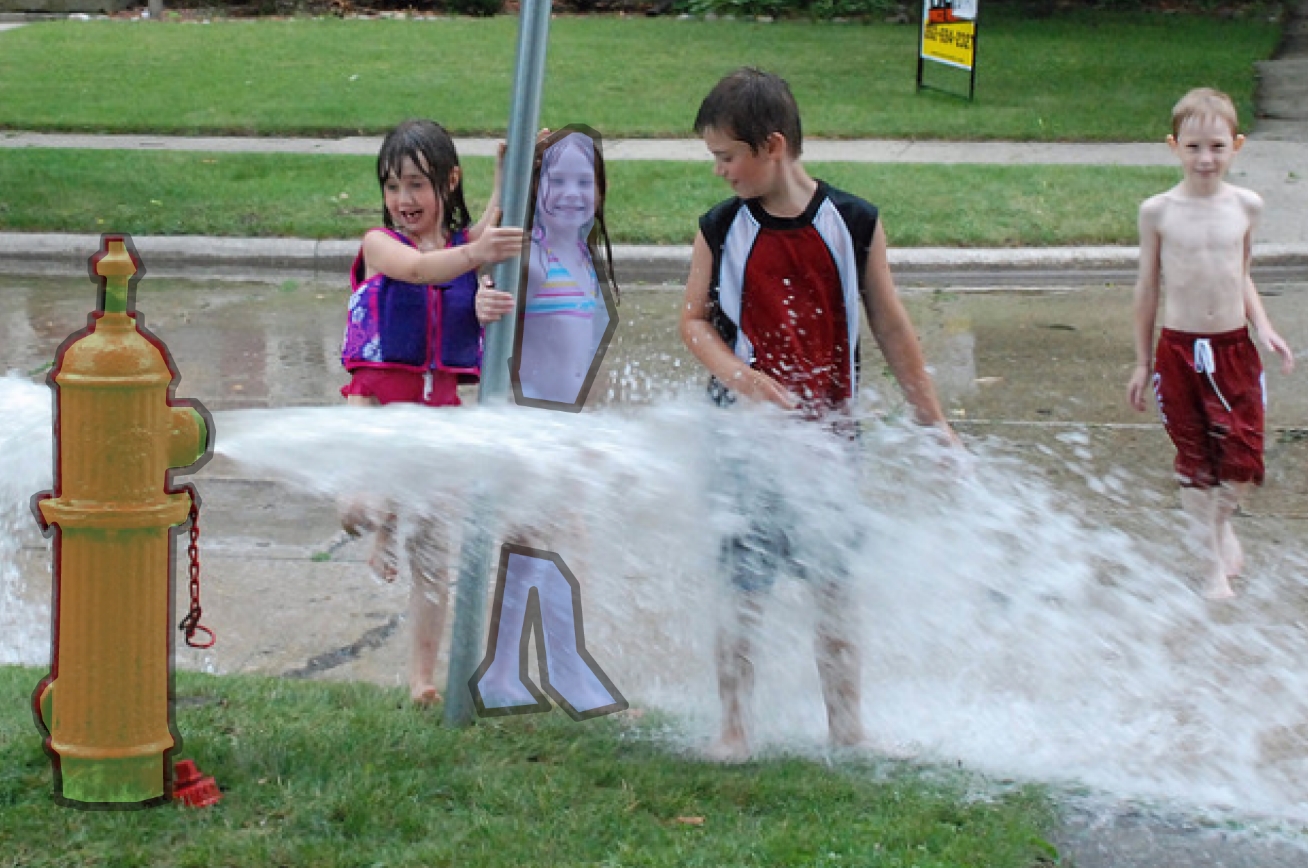}&
   \hspace*{-0.35cm}\includegraphics[height=0.23\textwidth]{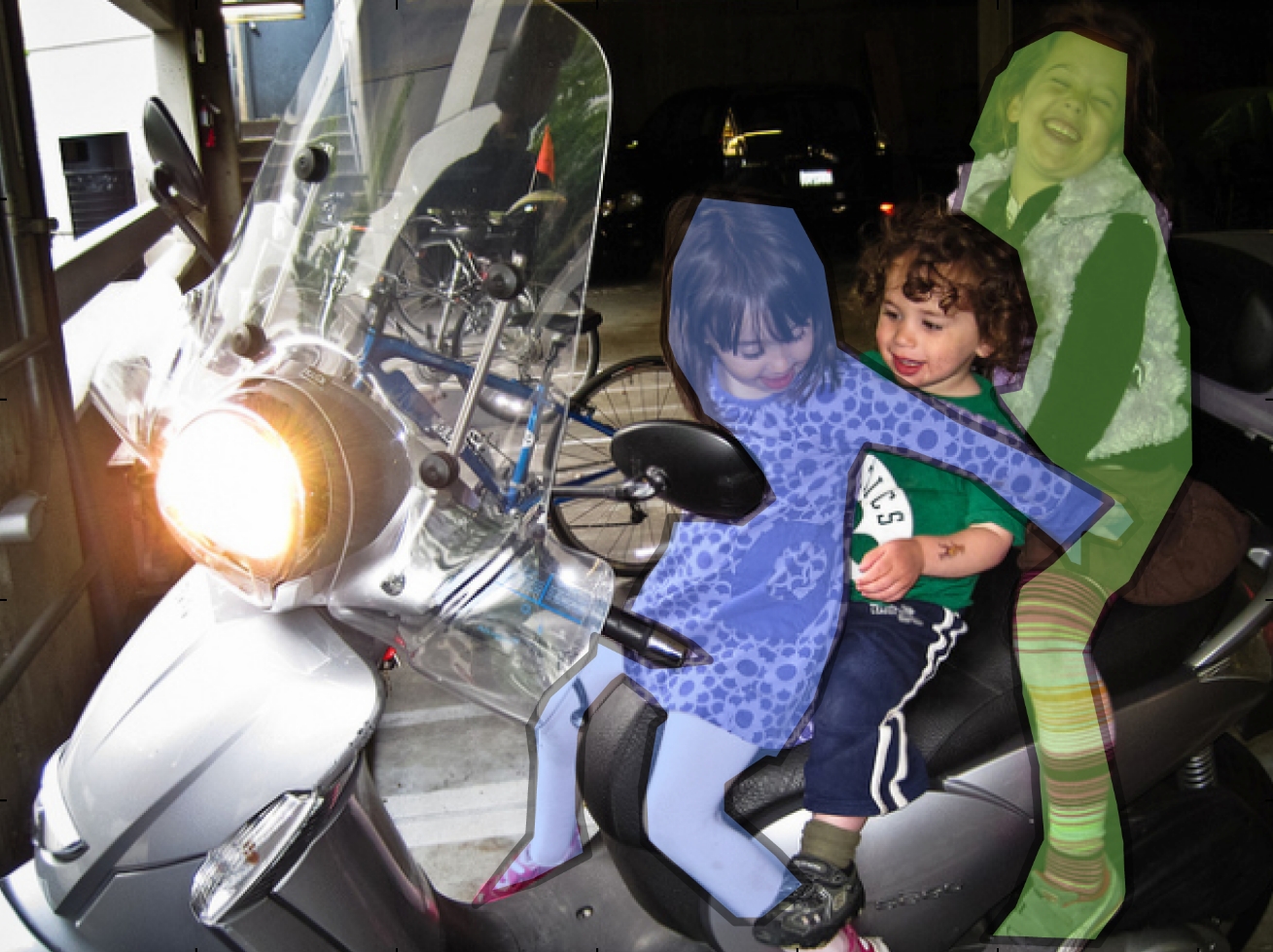}\\ 
   \hspace*{0.2cm}`happy' + `hydrant' & \hspace*{-0.35cm}`touch' + `behind'\\
\end{tabular}
\begin{tabular}{ccc}
   \hspace*{0.3cm}\includegraphics[height=0.23\textwidth]{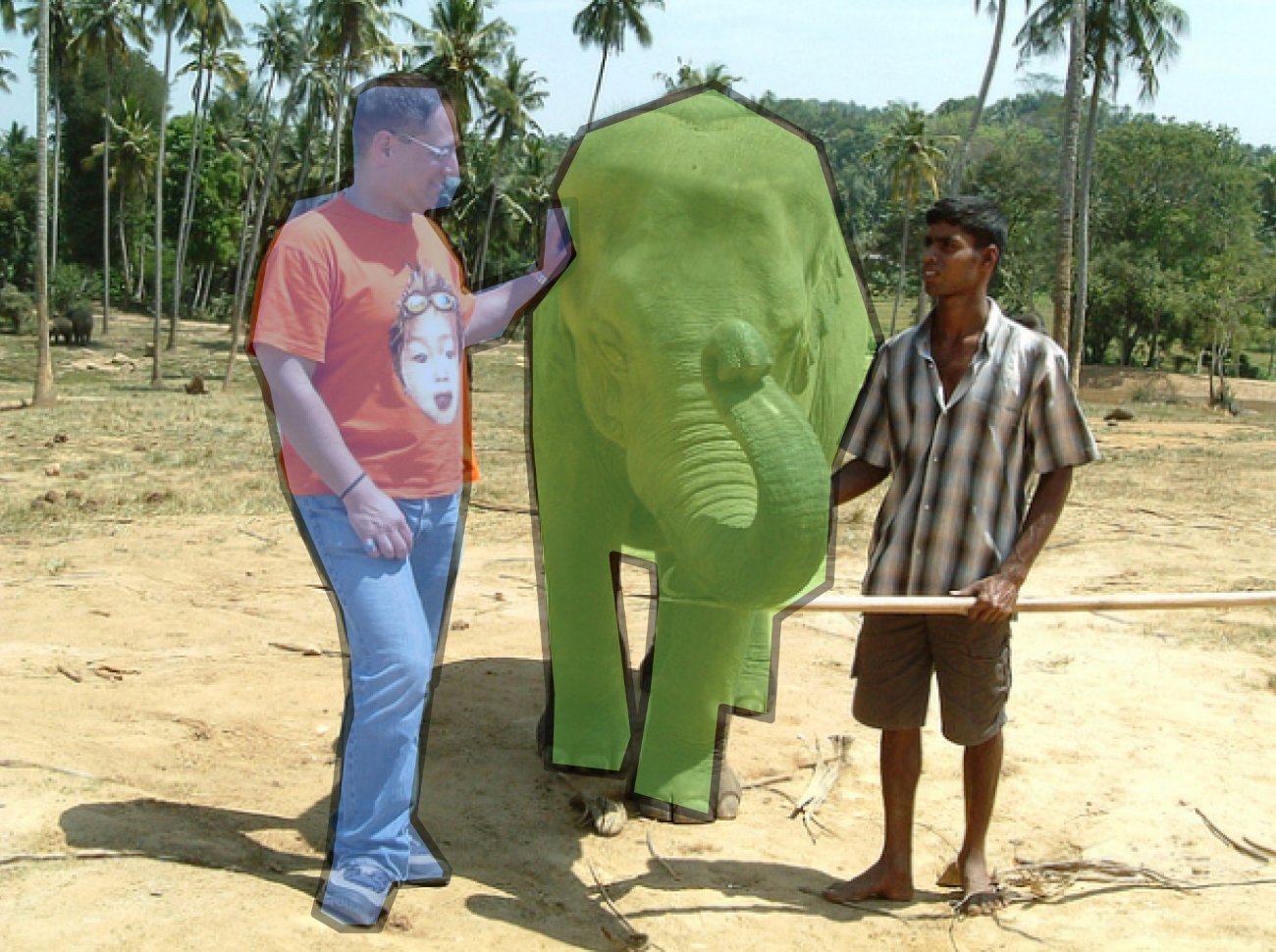}&
   \hspace*{-0.4cm}\includegraphics[height=0.23\textwidth]{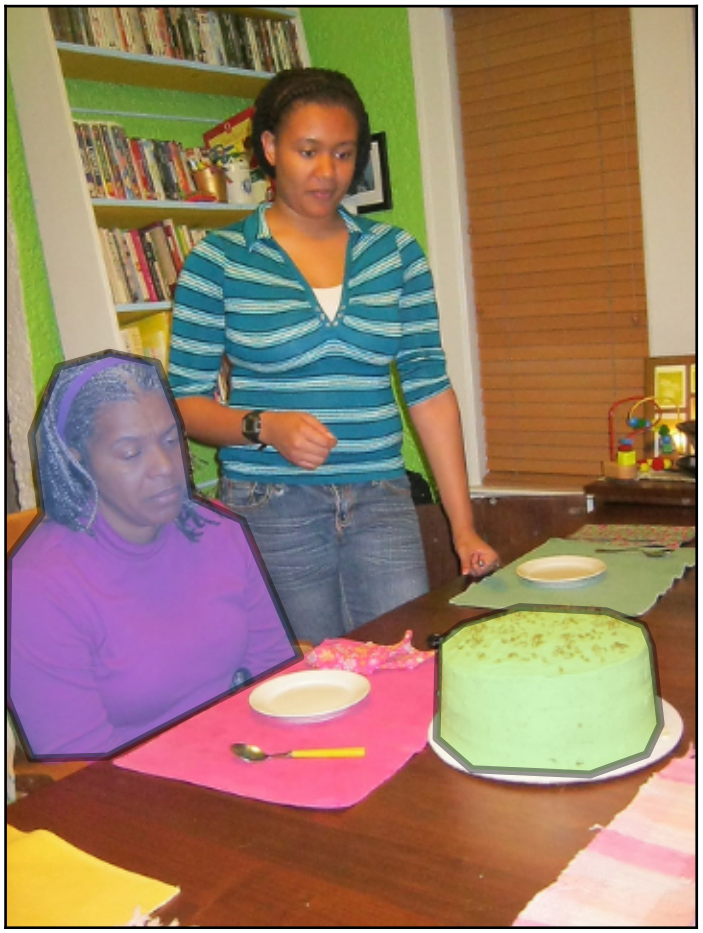}&
   \hspace*{-0.4cm}\includegraphics[height=0.23\textwidth]{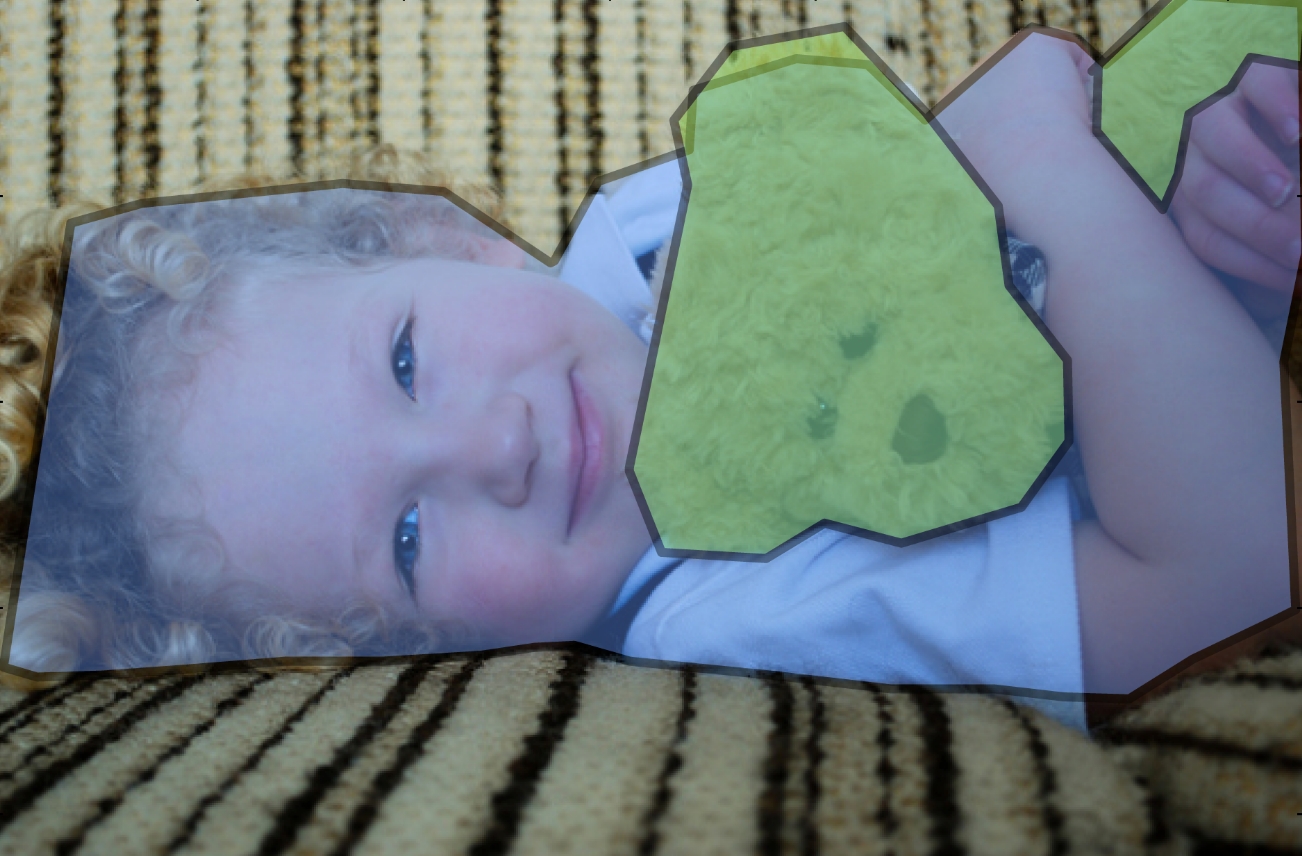}\\
   \hspace*{0.3cm}`happy' + `elephant' & \hspace*{-0.4cm}`sad' + `cake' & \hspace*{-0.4cm}`touch' + `above'\\
\end{tabular}
\end{center}
\vspace*{-0.5cm}
\caption{\textbf{Sample Query Results. } Sample images returned as a result of querying our dataset for visual actions with rare emotion, posture, position or location combinations. Subjects are highlighted in blue and the objects they are interacting with in green.}
\label{fig:example_queries}
\end{figure}

To explore the expressive power of our annotations we decided to query rare types of interactions and visualize the images retrieved. Figure~\ref{fig:example_queries} shows the result of querying our dataset for visual actions with rare emotion, posture, position or location combinations. The format of the annotations allows to query for images by specifying at the same time multiple properties of the interactions and their combinations, making them particularly suited for the training of image retrieval systems.

\newpage


\section{Discussion and Conclusions}

\vspace*{-0.13cm}

By a combined analysis of VerbNet and MS COCO captions we were able to compile a list of the main $140$ visual actions that take place in common scenes. Our list, which we call Visual VerbNet (VVN), attempts to include all actions that are visually discriminable. It avoids verb synonyms, actions that are specific to particular domains, and fine-grained actions. Unlike previous work, Visual VerbNet is not the result of experimenter's idiosyncratic choices; rather, it is derived from linguistic analysis (VerbNet) and an existing large dataset of everyday scenes (MS COCO captions).

Our novel dataset, COCO-a, consists of the VVN actions contained in $10,000$ MS COCO images. MS COCO images are representative of a wide variety of scenes and situations; 81 common objects are annotated in all images with pixel precision segmentations.  A key aspect of our annotations is that they are complete. First, each person in each image is identified as a possible subject, active agent of some action. Second, for each agent the set of objects that he/she is interacting with is identified. Third, for each agent-object pair (and each single agent) all the possible interactions involving that pair are identified, along with high level visual cues such as emotion and posture, spatial relationship and distance. The analysis of our annotations suggests that our collection of images ought to be augmented with an eye to increasing representation for the VVN actions that are less frequent in MS COCO.

We hope that our dataset will provide researchers with a starting point for conceptualizing about actions in images: which representations are most suitable, which algorithms should be used. We also hope that it will provide an ambitious benchmark on which to train and test algorithms. Amongst applications that are enabled by this dataset are building visual Q\&A systems~\cite{gemanTuringTest2015,DBLP:journals/corr/AntolALMBZP15}, more sophisticated image retrieval systems~\cite{johnson2015}, and automated analysis of actions in images of social media.

\clearpage
\bibliographystyle{amsplain}
\bibliography{arXiv_cocoa}

\providecommand{\bysame}{\leavevmode\hbox to3em{\hrulefill}\thinspace}
\providecommand{\MR}{\relax\ifhmode\unskip\space\fi MR }
\providecommand{\MRhref}[2]{%
  \href{http://www.ams.org/mathscinet-getitem?mr=#1}{#2}
}
\providecommand{\href}[2]{#2}
\begin{thebibliography}{10}

\bibitem{anderson2014toward}
David~J Anderson and Pietro Perona, \emph{Toward a science of computational
  ethology}, Neuron \textbf{84} (2014), no.~1, 18--31.

\bibitem{DBLP:journals/corr/AntolALMBZP15}
Stanislaw Antol, Aishwarya Agrawal, Jiasen Lu, Margaret Mitchell, Dhruv Batra,
  C.~Lawrence Zitnick, and Devi Parikh, \emph{{VQA:} visual question
  answering}, CoRR \textbf{abs/1505.00468} (2015).

\bibitem{bregler97}
C.~Bregler, \emph{Learning and recognizing human dynamics in video sequences},
  IEEE Conf. on Computer Vision and Pattern Recognition (CVPR), 1997,
  pp.~568--574.

\bibitem{deng2009imagenet}
Jia Deng, Wei Dong, Richard Socher, Li-Jia Li, Kai Li, and Li~Fei-Fei,
  \emph{Imagenet: A large-scale hierarchical image database}, Computer Vision
  and Pattern Recognition, 2009. CVPR 2009. IEEE Conference on, IEEE, 2009,
  pp.~248--255.

\bibitem{du2014compound}
Shichuan Du, Yong Tao, and Aleix~M Martinez, \emph{Compound facial expressions
  of emotion}, Proceedings of the National Academy of Sciences \textbf{111}
  (2014), no.~15, E1454--E1462.

\bibitem{ekman1992argument}
Paul Ekman, \emph{An argument for basic emotions}, Cognition \& emotion
  \textbf{6} (1992), no.~3-4, 169--200.

\bibitem{pascal-voc-2012}
M.~Everingham, L.~Van~Gool, C.~K.~I. Williams, J.~Winn, and A.~Zisserman,
  \emph{The {PASCAL} {V}isual {O}bject {C}lasses {C}hallenge 2012 {(VOC2012)}
  {R}esults},
  http://www.pascal-network.org/challenges/VOC/voc2012/workshop/index.html.

\bibitem{gemanTuringTest2015}
D.~Geman, S.~Geman, N.~Hallonquist, and L.~Younes, \emph{A visual turing test
  for computer vision system}, Proceedings of the National Academy of Sciences
  (PNAS) (2015).

\bibitem{guo2014survey}
Guodong Guo and Alice Lai, \emph{A survey on still image based human action
  recognition}, Pattern Recognition \textbf{47} (2014), no.~10, 3343--3361.

\bibitem{johnson2015}
J.~Johnson, R.~Krishna, M~Stark, Li-Jia Li, D.A. Shamma, M.S. Bernstein, and
  L.~Fei-Fei, \emph{Image retrieval using scene graphs}, IEEE Computer Vision
  and Pattern Recognition (CVPR), 2015.

\bibitem{kipper2008large}
Karin Kipper, Anna Korhonen, Neville Ryant, and Martha Palmer, \emph{A
  large-scale classification of english verbs}, Language Resources and
  Evaluation \textbf{42} (2008), no.~1, 21--40.

\bibitem{koller1991algorithmic}
D~Koller, N~Heinze, and HH~Nagel, \emph{Algorithmic characterization of vehicle
  trajectories from image sequences by motion verbs}, Computer Vision and
  Pattern Recognition, 1991. Proceedings CVPR'91., IEEE Computer Society
  Conference on, IEEE, 1991, pp.~90--95.

\bibitem{laptev2008learning}
Ivan Laptev, Marcin Marszalek, Cordelia Schmid, and Benjamin Rozenfeld,
  \emph{Learning realistic human actions from movies}, Computer Vision and
  Pattern Recognition, 2008. CVPR 2008. IEEE Conference on, IEEE, 2008,
  pp.~1--8.

\bibitem{le2014tuhoi}
Dieu-Thu Le, Jasper Uijlings, and Raffaella Bernardi, \emph{Tuhoi: Trento
  universal human object interaction dataset}, V\&L Net 2014 (2014), 17.

\bibitem{le2013exploiting}
Dieu-Thu Le, Jasper~RR Uijlings, and Raffaella Bernardi, \emph{Exploiting
  language models for visual recognition.}, EMNLP, 2013, pp.~769--779.

\bibitem{lin2014microsoft}
Tsung-Yi Lin, Michael Maire, Serge Belongie, James Hays, Pietro Perona, Deva
  Ramanan, Piotr Doll{\'a}r, and C~Lawrence Zitnick, \emph{Microsoft coco:
  Common objects in context}, Computer Vision--ECCV 2014, Springer, 2014,
  pp.~740--755.

\bibitem{nagel1988image}
H-H Nagel, \emph{From image sequences towards conceptual descriptions}, Image
  and vision computing \textbf{6} (1988), no.~2, 59--74.

\bibitem{nagel1994vision}
Hans-Hellmut Nagel, \emph{A vision of `vision and language' comprises action:
  An example from road traffic}, Artificial Intelligence Review \textbf{8}
  (1994), no.~2-3, 189--214.

\bibitem{ortony1990s}
Andrew Ortony and Terence~J Turner, \emph{What's basic about basic emotions?},
  Psychological review \textbf{97} (1990), no.~3, 315.

\bibitem{palmer99}
SE~Palmer, \emph{Vision science: Photons to phenomenology}, MIT Press, 1999.

\bibitem{polana1992recognition}
Ramprasad Polana and Randal~C Nelson, \emph{Recognition of motion from temporal
  texture}, Computer Vision and Pattern Recognition, 1992. Proceedings
  CVPR'92., 1992 IEEE Computer Society Conference on, IEEE, 1992, pp.~129--134.

\bibitem{rohr1994towards}
Karl Rohr, \emph{Towards model-based recognition of human movements in image
  sequences}, CVGIP: Image understanding \textbf{59} (1994), no.~1, 94--115.

\bibitem{russell1995modern}
Stuart Russell and Peter Norvig, \emph{Artificial intelligence, a modern
  approach}, Prentice-Hall, Egnlewood Cliffs, 1995.

\bibitem{schuldt2004recognizing}
Christian Schuldt, Ivan Laptev, and Barbara Caputo, \emph{Recognizing human
  actions: a local svm approach}, Pattern Recognition, 2004. ICPR 2004.
  Proceedings of the 17th International Conference on, vol.~3, IEEE, 2004,
  pp.~32--36.

\bibitem{von1988formalismus}
U~Cahn von Seelen, \emph{Ein formalismus zur beschreibung von bewegungsverben
  mit hilfe von trajektorien}, Ph.D. thesis, Diplomarbeit, Fakultaet fuer
  Informatik der Universitaet Karlsruhe, 1988.

\bibitem{yao2011human}
Bangpeng Yao, Xiaoye Jiang, Aditya Khosla, Andy~Lai Lin, Leonidas Guibas, and
  Li~Fei-Fei, \emph{Human action recognition by learning bases of action
  attributes and parts}, Computer Vision (ICCV), 2011 IEEE International
  Conference on, IEEE, 2011, pp.~1331--1338.

\end{thebibliography}

\clearpage

\section*{Appendix Overview}
In the appendix we provide:
\begin{enumerate}[(I)]
\item Statistics on the type of images in COCO-a.
\item Complete list of adverbs and visual actions.
\item Complete list of the objects of interactions and occurrence count.
\item Complete list of the visual actions and occurrence count.
\item User interface used to collect the interactions in the COCO-a dataset.
\item User interface used to collect the visual actions in the COCO-a dataset.
\end{enumerate}

\section*{Appendix I: Unbiased Nature of COCO-a}
\begin{figure}[h!]
  \begin{center}
\begin{tabular}{cc}
\includegraphics[width=0.25\textwidth]{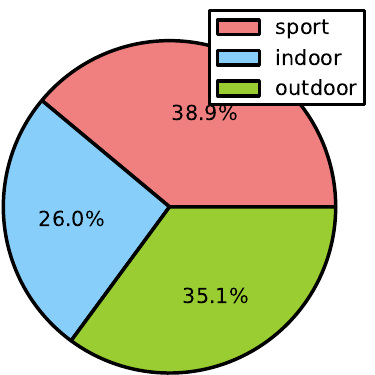} &
\includegraphics[width=0.25\textwidth]{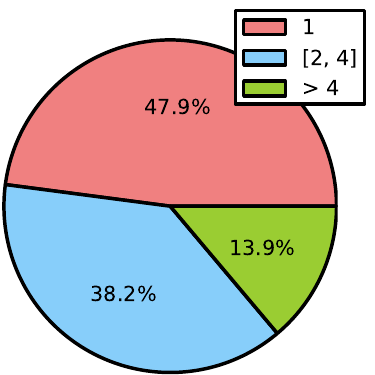}\\
\end{tabular}
  \end{center}
\caption{\textbf{Scene and subjects distributions.} (Left) The distribution of the type of scenes contained in the dataset. (Right) The distribution of the number of subjects appearing in each image.}
\label{fig:dataset-stats-images}
\end{figure}

We show in Figure~\ref{fig:dataset-stats-images} the unbiased nature of the images contained in our dataset. Different actions usually occur in different environments, so in order to balance the content of our dataset we selected an approximately equal number images of three types of scenes: sports, outdoors and indoors. We also selected images of various complexity, containing single subjects, small groups (2-4 subjects) and crowds ($>$4 subjects).

\section*{Appendix II: Visual Actions and Adverbs by Category}

In order to reduce the possibility of annotators using a term instead of another in the data collection interface, we organized visual actions into 8 groups -- `\textit{posture/motion}', `\textit{solo actions}', `\textit{contact actions}', `\textit{actions with objects}', `\textit{social actions}', `\textit{nutrition actions}', `\textit{communication actions}', `\textit{perception actions}'. This was based on two simple rules: (a) actions in the same group share some important property, e.g. being performed solo, with objects, with people, or indifferently with people and objects, or being an action of posture; (b) actions in the same group tend to be mutually exclusive, e.g. a person can be drinking or eating at a certain moment, not both. Furthermore, we included in our study 3 `adverb' categories: `\textit{emotion}' of the subject, `\textit{location}' and `\textit{relative distance}' of object with respect to the subject.

Tables~\ref{fig:full_metadata} and ~\ref{tab:full_visual_actions} contain a break down of the visual actions and adverbs into the categories that were presented to the Amazon Mechanical Turk workers.

\begin{table}[h!]
\begin{minipage}{\textwidth}
\begin{center}
\resizebox{0.7\textwidth}{!}{
\begin{tabular}[b]{|c|c|c|}
\hline
\multicolumn{3}{|c|}{\textbf{Adverbs}}\\
\hline
\begin{tabular}[b]{c}
Emotion (6)\\
\hline
anger\\
disgust\\
fear\\
happiness\\
sadness\\
surprise\\
\end{tabular}
&
\begin{tabular}[b]{c}
Relative Location (6)\\
\hline
above\\
behind\\
below\\
in front\\
left\\
right\\
\end{tabular}
&
\begin{tabular}[b]{c}
Relative Distance (4)\\
\hline
far\\
full contact\\
light contact\\
near\\
\\
\\
\end{tabular}\\
\hline
\end{tabular}
}
\end{center}
\end{minipage}
\medskip
\caption{\textbf{Adverbs ordered by category.} The complete list of high level visual cues collected, describing the subjects (emotion) and localization of the interaction (relative location and distance).}
\label{fig:full_metadata}
\end{table}

\begin{table}
\begin{minipage}{\textwidth}
\begin{center}

\resizebox{\textwidth}{!}{
\begin{tabular}[b]{|c|c|c|}
\hline
\multicolumn{3}{|c|}{\textbf{Visual Actions}}\\
\hline
\begin{tabular}[b]{ccc}
\multicolumn{3}{c}{Posture / Motion (23)}\\
\hline
balance & hang & run \\
bend & jump & sit \\
bow & kneel & squat \\
climb & lean & stand \\
crouch & lie & straddle \\
fall & perch & swim \\
float & recline & walk \\
fly & roll &  \\
\\
\\
\\
\end{tabular}
&
\begin{tabular}[b]{c}
Communication (6)\\
\hline
call\\
shout\\
signal\\
talk\\
whistle\\
wink\\
\\
\\
\\
\\
\\
\end{tabular}
&
\begin{tabular}[b]{cc}
\multicolumn{2}{c}{Contact (22)}\\
\hline
avoid & massage \\
bit & pet \\
bump & pinch \\
caress & poke \\
hit & pull \\
hold & punch \\
hug & push \\
kick & reach \\
kiss & slap \\
lick & squeeze \\
lift & tickle \\
\end{tabular}\\
\hline
\begin{tabular}[b]{ccc}
\multicolumn{3}{c}{Social (24)}\\
\hline
accompany & give  & play baseball   \\
be with   & groom & play basketball \\
chase     & help  & play frisbee    \\
dance     & hunt  & play soccer     \\
dine      & kill  & play tennis     \\
dress     & meet  & precede         \\
feed      & pay   &                 \\
fight     & shake hands &           \\
follow    & teach &                 \\
\end{tabular}
&
\begin{tabular}[b]{c}
Perception (5)\\
\hline
listen\\
look\\
sniff\\
taste\\
touch\\
\\
\\
\\
\\
\end{tabular}
&
\begin{tabular}[b]{c}
Nutrition (7)\\
\hline
chew\\
cook\\
devour\\
drink\\
eat\\
prepare\\
spread\\
\\
\\
\end{tabular}\\
\hline
\begin{tabular}[b]{cc}
\multicolumn{2}{c}{Solo (24)}\\
\hline
blow & play soccer \\
clap & play tennis \\
cry & play instrument \\
draw & pose \\
groan & sing \\
laugh & sleep \\
paint & smile \\
photograph & write \\
play & skate \\
play baseball & ski \\
play basketball & snowboard \\
play frisbee & surf \\
\end{tabular}
&
\multicolumn{2}{|c|}{
\begin{tabular}[b]{ccc}
\multicolumn{3}{c}{With objects (34)}\\
\hline
bend & fill & separate \\
break & get & show \\
brush & lay & spill \\
build & light & spray \\
carry & mix & steal \\
catch & pour & put \\
clear & read & throw \\
cut & remove & use \\
disassemble & repair & wash \\
drive & ride & wear\\
drop & row & \\
exchange & sail & \\
\end{tabular}
}\\
\hline
\end{tabular}
}
\end{center}
\end{minipage}
\medskip
\caption{\textbf{Visual actions ordered by category.} The complete list of visual actions contained in Visual VerbNet. Visual actions in one category are usually mutually exclusive, visual actions of different categories may co-occur.}
\label{tab:full_visual_actions}
\end{table}

\clearpage

\section*{Appendix III: Object Occurrences in Interactions}
\begin{figure}[h!]
\begin{minipage}[c]{\textwidth}
\centering
\begin{tabular}{c}
  \hspace*{-0.4cm} \includegraphics[width=\textwidth]{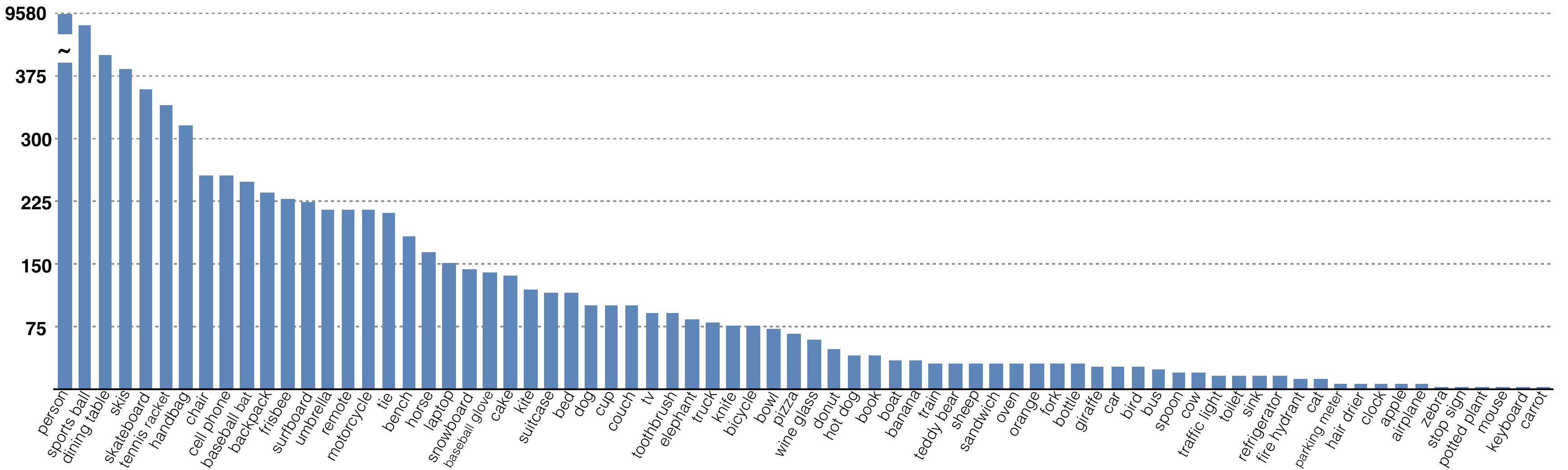}\\
\end{tabular}
\end{minipage}
\vspace*{-0.5cm}
\caption{\textbf{Most frequent objects.} The complete lists of interacting objects obtained from the annotators. The scale is linear.}
\label{fig:full_object_list}
\end{figure}

We show the  full lists of objects that people interact in Figure~\ref{fig:full_object_list}.

\section*{Appendix IV: Visual Action Occurrences}
\begin{figure}[h!]
\begin{minipage}[c]{\textwidth}
\centering
\begin{tabular}{c} 
  \hspace*{-0.4cm} \includegraphics[width=\textwidth]{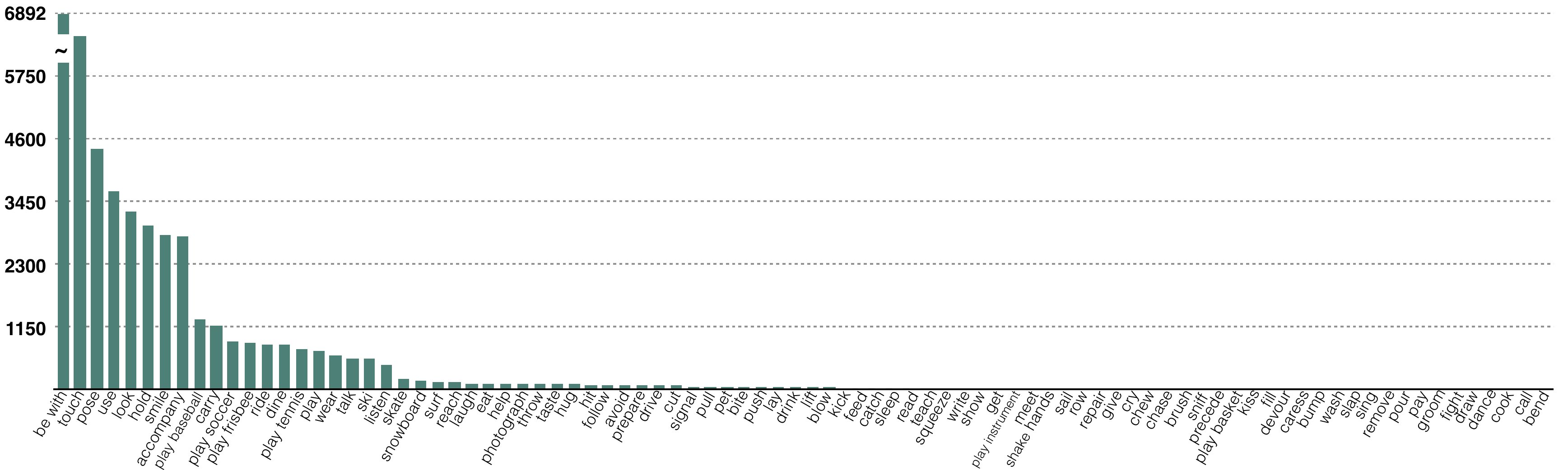}\\
\end{tabular}
\end{minipage}
\vspace*{-0.5cm}
\caption{\textbf{Most frequent visual actions. }The complete lists of `visual actions' obtained from the annotators. The scale is linear. }
\label{fig:full_verb_list}

\vspace*{-0.0cm}
\centering
\includegraphics[width=0.8\textwidth]{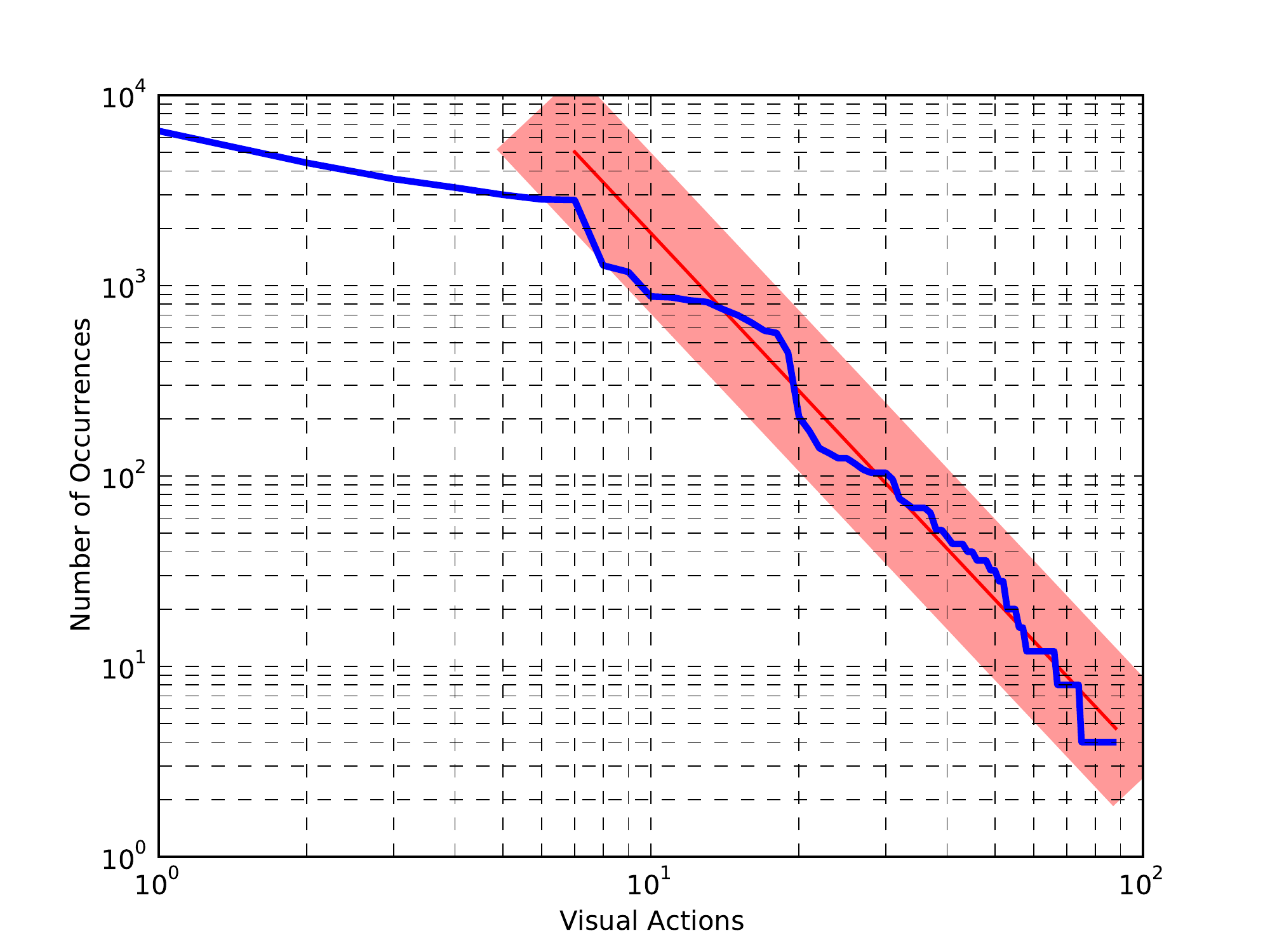}
  \vspace*{-0.5cm}
\caption{\textbf{Visual actions heavy tail analysis. }The plot in log-log scale of the list of visual actions against the number of occurrences. See also Fig.\ref{fig:object_verb_list}.}
\label{fig:verbs-list-tail}
\end{figure}

We show the complete lists of `visual actions' annotated from the images and their occurrences in Figure~\ref{fig:full_verb_list}.

If we consider tail all the actions with less than $2000$ occurrences then $90\%$ of the actions are in the tail and cover $27\%$ of the total number of occurrences. The distribution of the visual actions' counts follows a heavy tail distribution, to which we fit a line, shown in Figure~\ref{fig:verbs-list-tail}, with slope $\alpha \sim -3$. This seems to indicate that the MS COCO dataset is sufficient for a thorough representation and study of about 20 to 30 visual actions, however we are considering methods to bias our image selection process in order to obtain more samples of the actions contained in the tail.

\section*{Appendix V: Interactions User Interface}

\begin{figure}[h!]
\centering
\begin{minipage}[c]{\textwidth}
\includegraphics[width=\textwidth]{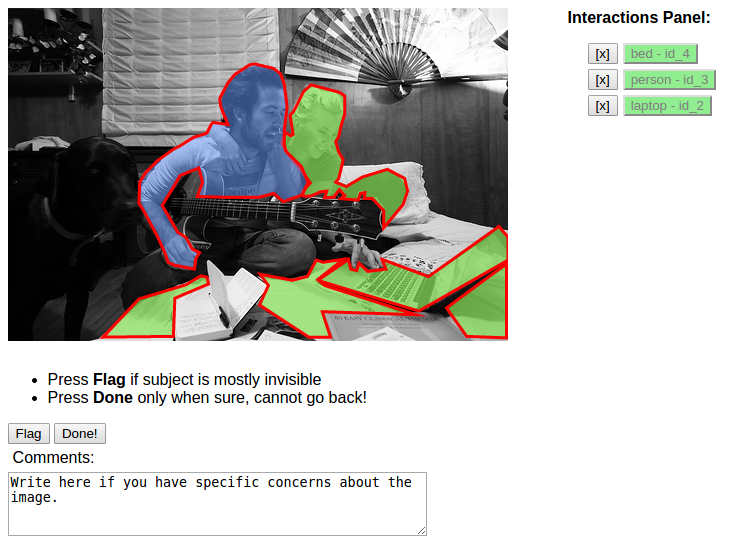}
\end{minipage}
\caption{\textbf{Interactions GUI.} In this image the blue subject is interacting with another person, the bed and the laptop.}
\label{fig:gui-interactions}
\end{figure}

In Figure~\ref{fig:gui-interactions} we show the AMT interface developed to collect interaction annotations from images. Each worker is presented with a series of 10 images, each containing a subject highlighted in blue and asked to (1) flag the subject if it is mostly occluded or invisible; (2) if the subject is sufficiently visible, click on all the objects he/she is interacting with. The interface provides feedback to the annotator by highlighting in white all the annotated objects when the mouse is hovered over the image, and selecting in green the objects once they are clicked. Annotators can remove annotations by either clicking on the object segmentation on the image a second time or using the appropriate button in the annotation panel.
We included a comments text box to obtain specific feedback workers on each image.

\clearpage

\section*{Appendix VI: Visual Actions User Interface}

\begin{figure}[h!]
\begin{minipage}[c]{\textwidth}
\centering
\begin{tabular}{c}
\includegraphics[width=0.4\textwidth]{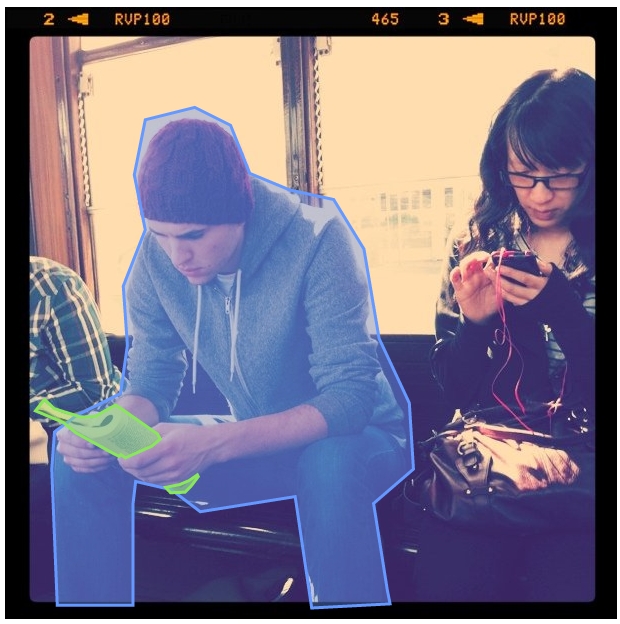}\\
Step 1: Flag the interaction if subject is occluded \\
\includegraphics[width=\textwidth]{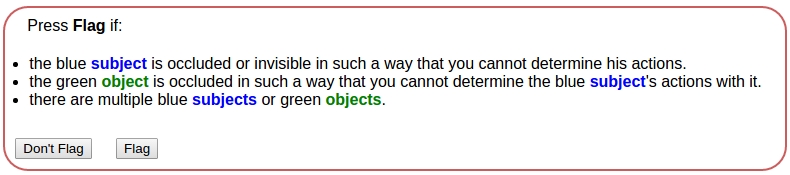}\\
Step 2: Provide Relative Location\\
\includegraphics[width=\textwidth]{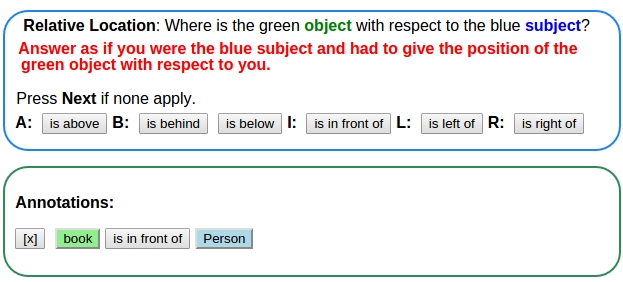} \\
\end{tabular}
\end{minipage}
\caption{\textbf{Visual Actions GUI.} }
\label{fig:gui-verbs1}
\end{figure}

\begin{figure}[h!]
\begin{minipage}[c]{\textwidth}
\centering
\begin{tabular}{c}
Step 3: Provide Distance of Interaction \\
\includegraphics[width=\textwidth]{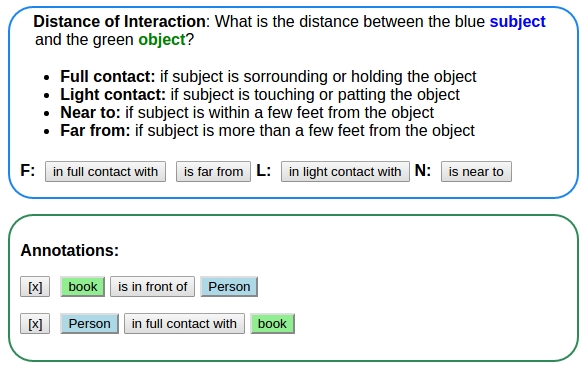} \\
Step 4: Provide Senses used in Interaction \\
\includegraphics[width=\textwidth]{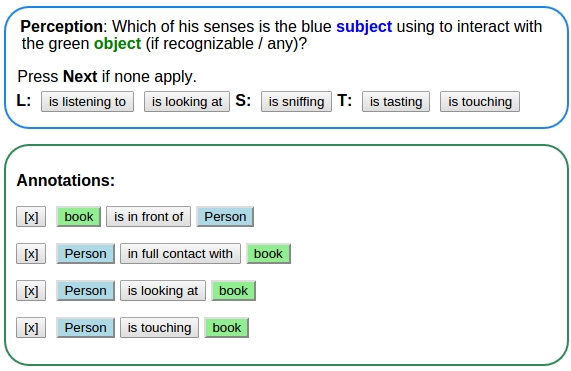} \\
Step 5: Provide Nutrition Visual Actions (none in this case) \\
\includegraphics[width=\textwidth]{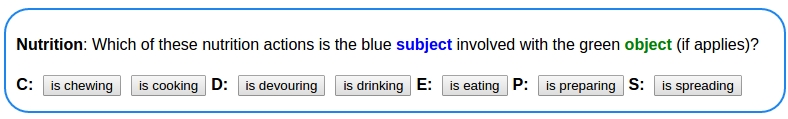} \\
\end{tabular}
\end{minipage}
\caption{\textbf{Visual Actions GUI.} }
\label{fig:gui-verbs2}
\end{figure}

\begin{figure}[t!]
\centering
\begin{minipage}[c]{\textwidth}
\begin{tabular}[c]{c}
Step 6: Provide Contact Visual Actions (free-typing is allowed)\\
\includegraphics[width=\textwidth]{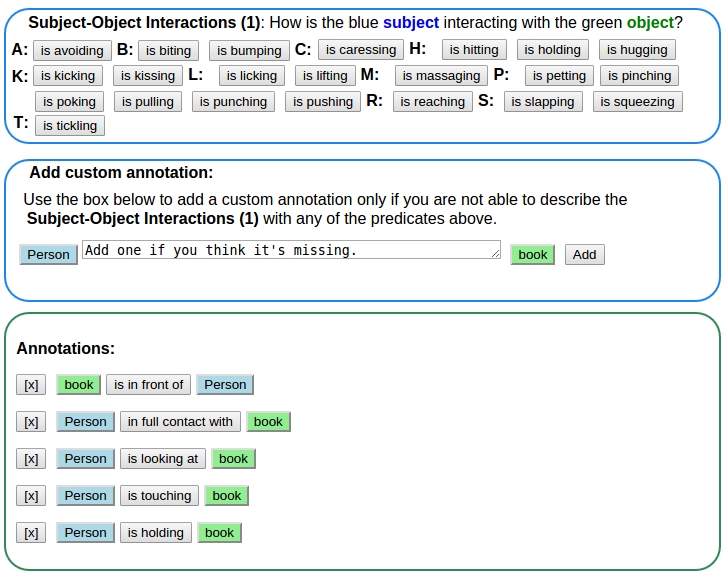} \\
Step 7: Provide Object Visual Actions (free-typing is allowed)\\
\includegraphics[width=\textwidth]{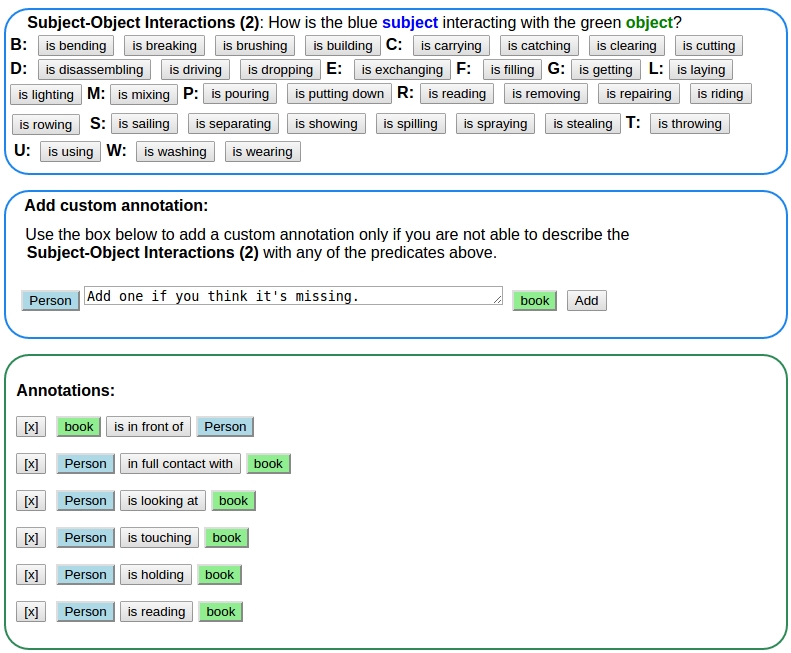} \\
\end{tabular}
\end{minipage}
\caption{\textbf{Visual Actions GUI.} }
\label{fig:gui-verbs3}
\end{figure}

In Figures~\ref{fig:gui-verbs1},~\ref{fig:gui-verbs2} and~\ref{fig:gui-verbs3} we show the sequences of steps required in the AMT interface developed to collect visual action annotations. We collect visual actions for all the interactions obtained from the previously shown GUI having an agreement of 3 out of 5 workers, as explained in more details in Section~\ref{sec:interactions_anno}.
Each worker is presented with a single image containing a subject (highlighted in blue) and an object (highlighted in green) and asked to go through 8 panels, one for each category of visual actions, and select all the visual actions that apply to the visualized interaction. Annotators can skip a category if no visual action applies (i.e. nutrition visual actions only apply for food items). As they proceed through the 8 panels workers have the chance to visualize all the annotations that are being provided for the specific interaction, which helps avoid ambiguous annotations.
Depending on the object involved in the interaction some panels might not be shown (i.e. the \textit{communication} panel is not shown when the object of interaction is inanimate, as well as the \textit{nutrition} panel is not shown when the object of interaction is another person).


\end{document}